\documentclass{article}

\usepackage[preprint,nonatbib]{neurips_2023}

\usepackage[utf8]{inputenc} 
\usepackage[T1]{fontenc}    
\usepackage{hyperref}       
\usepackage{amsfonts}       
\usepackage{nicefrac}       
\usepackage{xcolor}         
\usepackage[nolist]{acronym}
\usepackage{algorithm,algorithmic}
\usepackage{amsmath,amsfonts}
\usepackage{array}
\usepackage[caption=false,font=normalsize,labelfont=sf,textfont=sf]{subfig}
\usepackage{textcomp}
\usepackage{stfloats}
\usepackage{url}
\usepackage{verbatim}
\usepackage{graphicx}
\usepackage{cite}
\usepackage{paralist}
\usepackage[font=small,labelfont=bf]{caption}

\usepackage[utf8]{inputenc} 
\usepackage{amsmath,dsfont,bbm,epsfig,amssymb,amsfonts,amstext,verbatim,amsopn}
\usepackage{multirow,multicol,lipsum,xfrac}
\usepackage{amsthm}
\usepackage{mathtools,amsthm}
\usepackage{url}
\usepackage{amsfonts}
\usepackage{tikz,pgfplots}
\usepackage[nolist]{acronym}
\usepgfplotslibrary{fillbetween}

\newcommand{\setL}{\mathbbmss{L}}

\newcommand{\setP}{\mathbbmss{P}}
\newcommand{\setR}{\mathbbmss{R}}

\newcommand{\setS}{\mathbbmss{S}}

\newcommand{\setT}{\mathbbmss{T}}

\newcommand{\setD}{\mathbbmss{D}}
\newcommand{\setF}{\mathbbmss{F}}

\newcommand{\Ex}[2]{ \mathbbm{E}_{#2} \left\lbrace #1 \right\rbrace }

\newcommand{\id}[2]{ _{ #2 \left[ #1\right]} }

\newcommand{\rmg}{\mathrm{g}}

\newcommand{\argmaxk}{\mathop{\mathrm{argmax}^k}}

\newcommand{\man}{\mathcal{N}}

\newcommand{\mal}{\mathcal{L}}

\newcommand{\bxx}{\mathbf{x}}

\newcommand{\bss}{\mathbf{s}}

\newcommand{\byy}{\mathbf{y}}

\newcommand{\bg}{{\mathbf{g}}}
\newcommand{\btheta}{{\boldsymbol{\theta}}}
\newcommand{\bdelta}{{\boldsymbol{\Delta}}}

\newcommand{\beps}{{\boldsymbol{\varepsilon}}}

\newcommand{\bxi}{{\boldsymbol{\xi}}}

\newcommand{\bx}{{\boldsymbol{x}}}

\newcommand{\set}[1]{\left\lbrace#1\right\rbrace}

\newcommand{\brc}[1]{\left( #1 \right) }
\newcommand{\inner}[1]{\left\langle #1 \right\rangle }

\newcommand{\dbc}[1]{\left[ #1 \right] }

\newcommand{\bt}{{\mathbf{t}}}

\newcommand{\bzz}{{\mathbf{z}}}
\newcommand{\baa}{{\mathbf{a}}}

\newcommand{\dif}{\mathrm{d}}

\newcommand{\by}{{\boldsymbol{y}}}

\newcommand{\be}{{\mathbf{e}}}

\newcommand{\trp}{\mathsf{T}}

\newcommand{\mI}{\mathbf{I}}

\newcommand{\mX}{\mathbf{X}}

\newcommand{\Top}[2]{\mathrm{Top}_{#1} \left( #2 \right)}

\newcommand{\norm}[1]{\left\lVert #1 \right\rVert}

\newcommand{\abs}[1]{\lvert #1 \rvert}

\newtheoremstyle{mystyle}
{}
{}
{\it}
{}
{\bfseries}
{:}
{ }
{}

\theoremstyle{mystyle}

\newtheorem{definition}{Definition}
\newtheorem{proposition}{Proposition}
\newtheorem{remark}{Remark}

\newtheorem{result}{Result}
\newtheorem{lemma}{Lemma}

\newcounter{bar}

\usepackage{xspace}
\newcommand{\tpK}{\textsc{Top}-$k$\xspace}
\newcommand{\rgtpK}{\textsc{RegTop}-$k$\xspace}
\usetikzlibrary{calc}

\makeatletter
\def\thanks#1{\protected@xdef\@thanks{\@thanks
		\protect\footnotetext{#1}}}
\makeatother

\title{Regularized Top-$k$: A Bayesian Framework for Gradient Sparsification} 

\author{%
	Ali Bereyhi \\
	University of Toronto\\
	\texttt{ali.bereyhi@utoronto.ca} 
	\thanks{This paper has been published in IEEE Transactions on Signal Processing, vol. 73, pp. 4463 - 4478, 2025. DOI: 10.1109/TSP.2025.3624791. The present arXiv version contains additional experimental results.}
	\thanks{This work was presented in part at the 34th IEEE International Workshop on Machine Learning for Signal Processing (MLSP) in September 2024 \cite{MLSP2024bereyhi}.}
	\thanks{This work was funded in part by Ericsson Canada and by the Natural Sciences and Engineering Research Council of Canada.}
	\thanks{All codes are available at \href{https://github.com/AlBerey/RegTopK}{\texttt{https://github.com/AlBerey/RegTopK}}.}
	\And
	Ben Liang \\
	University of Toronto\\
	\texttt{liang@utoronto.ca}
	\And
	Gary Boudreau\\
	Ericsson Inc Canada\\
	\texttt{gary.boudreau@ericsson.com}
	\And
	Ali Afana \\
	Ericsson Inc Canada\\
	\texttt{ali.afana@ericsson.com}
}
\begin{document}
	\maketitle

\begin{acronym}
	\acro{mimo}[MIMO]{multiple-input multiple-output}
	\acro{simo}[SIMO]{single-input multiple-output}
	\acro{csi}[CSI]{channel state information}
	\acro{awgn}[AWGN]{additive white Gaussian noise}
	\acro{iid}[i.i.d.]{independent and identically distributed}
	\acro{uts}[UTs]{user terminals}
	\acro{ps}[PS]{parameter server}
	\acro{irs}[IRS]{intelligent reflecting surface}
	\acro{tas}[TAS]{transmit antenna selection}
	\acro{glse}[GLSE]{generalized least square error}
	\acro{rhs}[r.h.s.]{right hand side}
	\acro{lhs}[l.h.s.]{left hand side}
	\acro{wrt}[w.r.t.]{with respect to}
	\acro{rs}[RS]{replica symmetry}
	\acro{mac}[MAC]{multiple access channel}
	\acro{np}[NP]{non-deterministic polynomial-time}
	\acro{papr}[PAPR]{peak-to-average power ratio}
	\acro{rzf}[RZF]{regularized zero forcing}
	\acro{snr}[SNR]{signal-to-noise ratio}
	\acro{sinr}[SINR]{signal-to-interference-and-noise ratio}
	\acro{svd}[SVD]{singular value decomposition}
	\acro{mf}[MF]{matched filtering}
	\acro{gamp}[GAMP]{generalized AMP}
	\acro{amp}[AMP]{approximate message passing}
	\acro{vamp}[VAMP]{vector AMP}
	\acro{map}[MAP]{maximum-a-posterior}
	\acro{ml}[ML]{maximum likelihood}
	\acro{mse}[MSE]{mean squared error}
	\acro{mmse}[MMSE]{minimum mean squared error}
	\acro{ap}[AP]{average power}
	\acro{ldgm}[LDGM]{low density generator matrix}
	\acro{tdd}[TDD]{time division duplexing}
	\acro{rss}[RSS]{residual sum of squares}
	\acro{rls}[RLS]{regularized least-squares}
	\acro{ls}[LS]{least-squares}
	\acro{erp}[ERP]{encryption redundancy parameter}
	\acro{zf}[ZF]{zero forcing}
	\acro{ta}[TA]{transmit-array}
	\acro{ofdm}[OFDM]{orthogonal frequency division multiplexing}
	\acro{dc}[DC]{difference of convex}
	\acro{bcd}[BCD]{block coordinate descent}
	\acro{mm}[MM]{majorization-maximization}
	\acro{bs}[BS]{base-station}
	\acro{aircomp}[AirComp]{over-the-air-computation}
	\acro{ULA}[ULA]{uniform linear array}
	\acro{fl}[FL]{federated learning}
	\acro{fedavg}[FedAvg]{federated averaging}
	\acro{ota-fl}[OTA-FL]{over-the-air federated learning}
	\acro{los}[LoS]{line-of-sight}
	\acro{nlos}[NLoS]{non-line-of-sight}
	\acro{aoa}[AoA]{angle of arrival}
	\acro{nn}[NN]{neural network}
	\acro{cnn}[CNN]{convolutional neural network}
	\acro{dnn}[DNN]{deep neural network}
	\acro{sgd}[SGD]{stochastic gradient descent}
	\acro{aircomp}[AirComp]{over-the-air computation}
	\acro{lmi}[LMI]{linear matrix inequalities}
	\acro{qcqp}[QCQP]{quadratically constrained quadratic programming}
	\acro{lln}[LLN]{law of large numbers}
	\acro{clt}[CLT]{central limit theorem}
	\acro{gs}[GS]{gradient sparsification}
	\acro{ota-f}[OTA-F]{over-the-air fair}
	\acro{bc}[BC]{broadcast channel}
	\acro{pdf}[PDF]{probability density function}
\end{acronym}

	\begin{abstract}
	Error accumulation is effective for gradient sparsification in distributed settings: initially-unselected gradient entries are eventually selected as their accumulated error exceeds a certain level. The accumulation essentially behaves as a scaling of the learning rate for the selected entries. Although this property prevents the slow-down of lateral movements in distributed gradient descent, it can deteriorate convergence in some settings. This work proposes a novel sparsification scheme that controls the learning rate scaling of error accumulation. The development of this scheme follows two major steps: first, gradient sparsification is formulated as an inverse probability (inference) problem, and the Bayesian optimal sparsification mask is derived as a  maximum-a-posteriori estimator. Using the prior distribution inherited from \tpK, we derive a new sparsification algorithm which can be interpreted as a regularized form of \tpK. We call this algorithm \textit{regularized} \tpK (\textsc{RegTop-}$k$). It utilizes past aggregated gradients to evaluate posterior statistics of the next aggregation. It then prioritizes the local accumulated gradient entries based on these posterior statistics. We validate our derivation through various numerical experiments. In distributed linear regression, it is observed that while \tpK remains at a fixed distance from the global optimum, \rgtpK converges to the global optimum at significantly higher compression ratios. We further demonstrate the generalization of this observation by employing \rgtpK in distributed training of ResNet-18 on CIFAR-10, as well as fine-tuning of multiple computer vision models on the ImageNette dataset. Our numerical results confirm that as the compression ratio increases, \rgtpK sparsification noticeably outperforms \tpK. 
\end{abstract}



\section{Introduction}
\label{sec:intro}
Consider the standard setting for distributed \ac{sgd} \cite{verbraeken2020survey}, where $N$ workers compute \textit{local} gradients and share them with a server to estimate the \textit{global} gradient. In iteration $t$, worker $n$ starts from a common model $\btheta^t\in\setR^J$ and determines its local gradient using a local surrogate loss function that is determined by averaging a loss over a stochastically-selected mini-batch. We denote this local gradient by $\bg_n^{t}\in\setR^J$. After receiving $\bg_1^{t}, \ldots, \bg_N^t$, the server estimates the global gradient $\bg^t$ by weighted averaging, i.e., %
\begin{align}
	\bg^{t} =   \sum_{n=1}^N \omega_n  \bg_n^{t} %
	\label{eq:glob}
\end{align}
for some weights $\omega_n$, which are often proportional to the local batch sizes. This estimate is then sent back to the workers to update $\btheta^{t+1} = \btheta^t - \eta^t \bg^{t}$ with a learning rate $\eta^t$. 

With realistic data networks, the scale of communication in the above setting can be prohibitive. For instance in ResNet-110 \cite{he2016deep}, $J\approx1.7 \times 10^6$. Assuming $1000$ mini-batches at each worker, the network exchanges $1.7 \times 10^9$ symbols per epoch for each worker.  %
Various approaches have been developed in the literature to address this challenge through improving the communication-efficiency of distributed learning algorithms. 
This work focuses on \textit{gradient sparsification}, which is a well-known approach for model compression in distributed \ac{sgd}. 

\subsection{Communication-Efficient Distributed Learning}
A basic approach to improve communication efficiency is to minimize the number of communication rounds. This scheme suggests that each worker performs multiple local updates in each round; see for instance the classical \ac{fedavg} algorithm \cite{mcmahan2017communication}.  The communication reduction in this approach comes at the expense of drift at the converging model: as the number of local updates increases, the local gradients start to drift towards the local optima causing the global model to converge to a point  away from the global optimum 
\cite{charles2021convergence,malinovskiy2020local}. 
Several studies have proposed approaches to suppress this drift; among them are Scaffold \cite{karimireddy2020scaffold}, FedSplit \cite{pathak2020fedsplit}, FedProx \cite{li2020federated}, FedNova \cite{wang2020tackling} and FedLin \cite{mitra2021linear}. In a nutshell, these lines of work try to compensate the impact of local drift by either controlling local learning rates or modifying the local updates, e.g., Scaffold and FedLin. 

An alternative approach is to communicate low-resolution quantization of local gradients that in the extreme case can carry only the sign, i.e., one-bit quantization
\cite{seide20141,grubic2018synchronous}. Various quantization approaches are developed and investigated in the literature; see for instance 
\cite{alistarh2017qsgd,gandikota2021vqsgd,sun2019communication,abdi2020quantized}. Further communication reduction can be achieved in this approach by compressing the quantized gradients via \textit{unbiased} estimators \cite{abdi2020quantized}.

This work focuses on a third class of schemes for improving communication-efficiency, namely model compression. In its general form, model compression suggests to compress the exchanged parameters with a potentially lossy technique, e.g., sketching \cite{ivkin2019communication,rothchild2020fetchsgd,zhang2024sketch} or sparsification  \cite{strom2015scalable,wen2017terngrad}. This approach is particularly effective in settings with significant redundancy in local parameters. A classical model compression scheme is \textit{gradient sparsification}, which is widely employed in implementation of distributed \ac{sgd} \cite{strom2015scalable,alistarh2018convergence}: in each iteration, the workers send only \textit{important gradient entries} along with their indices. The term \textit{sparsification} refers to the fact that the server treats the missing entries as zero, which can alternatively be interpreted as if the sever approximates the local gradients with their \textit{sparsified version}. Typical sparsity range used in practice is less than $1\%$ \cite{strom2015scalable,dryden2016communication,aji2017sparse,lin2018deep,sahu2021rethinking}. 

\subsection{Gradient Sparsification via \tpK}
The key point in the design of a gradient sparsification algorithm is to develop a mechanism that finds \textit{important gradient entries}. This can be challenging, as each worker has no explicit information about the gradients of the other workers, and hence can only decide \textit{locally}. A naive approach for sparsification is to select the largest local gradient entries in each iteration. Nevertheless, such approach can easily deteriorate the convergence of distributed \ac{sgd}; see the toy example given in Section~\ref{sec:example}. To overcome this issue, the \tpK sparsification algorithm invokes the idea of \textit{error accumulation} to keep track of those gradient entries that have been ignored in previous iterations \cite{aji2017sparse}.  


\tpK selects the $k$ largest \textit{accumulated} gradient entries in each iteration. Worker $n$ initializes its sparsification error with $\beps_n^0 = \boldsymbol{0}$. In iteration $t$, it computes its local gradient $\bg_n^t$ and the accumulated gradient as $\baa_n^t = \beps_n^{t} + \bg_n^t$. It then selects the $k$ entries of $\baa_n^t$ with the largest amplitude which results in the sparsified gradient $\hat{\bg}_n^t \in\setR^J$.  The sparsification error is then updated as $\beps_n^t = \baa_n^t - \hat{\bg}_n^t$. Through this error accumulation mechanism, the initially unselected entries get the chance of being selected after their errors become large enough. Then, \ac{sgd} moves a large step in the selected direction whose length is proportional to the \textit{accumulated} gradient entry. This behavior is known as \textit{learning rate scaling}, which significantly improves the convergence of sparsified distributed \ac{sgd} \cite{lin2018deep}. 

Though learning rate scaling is fairly effective for smooth loss functions, it can result in either alternation around the optimum or divergence for other loss functions. The error accumulation is equivalent to postponing the move in a particular direction to later iterations and then applying a \textit{scaled learning rate}. This can lead to convergence issues when the scaling is excessive. Considering this challenge, most studies restrict their analyses to the assumption that such deteriorating scaling does not occur in the network, e.g., see Assumption~1 in \cite{alistarh2018convergence}. 

In this work, we propose the \rgtpK algorithm for sparsification, which controls the learning rate scaling property.  
Our motivation for the development of \rgtpK is best understood through a toy example that is given next.
%

\subsection{Motivational Toy Example}
\label{sec:example}
Consider a simple logistic regression problem with $J=2$, where two workers employ distributed gradient descent to minimize the cross-entropy. Let workers 1 and 2 have single data-points $\brc{\bxx_1, 1}$ and $\brc{\bxx_2, 1}$, respectively, where $\bxx_1 = \dbc{100,1},$ and $\bxx_2 = \dbc{-100,1}$. The workers train a logistic regression model with weight vector $\btheta = \dbc{\theta_1, \theta_2}$  and zero bias. The loss of worker $n\in\set{1,2}$ in this case is given by
\begin{align}
	F_n \brc{\btheta} &=  \log \brc{1+ \exp\set{- \inner{\btheta;\bxx_n} }},
\end{align}
and the empirical risk used for distributed training is  
\begin{align}
F \brc{\btheta} =\frac{F_1 \brc{\btheta} + F_2 \brc{\btheta}}{2}.
\end{align}
The local gradient at worker $n$ is given by
\begin{align}
	\bg_n = -\frac{  \exp\set{- \inner{\btheta;\bxx_n} }  \bxx_n }{ 1+ \exp\set{- \inner{\btheta;\bxx_n} } },
\end{align}
and the global gradient is $\bg = 0.5\brc{\bg_1 + \bg_2} $. 

\begin{figure}
	\begin{center}
		\input{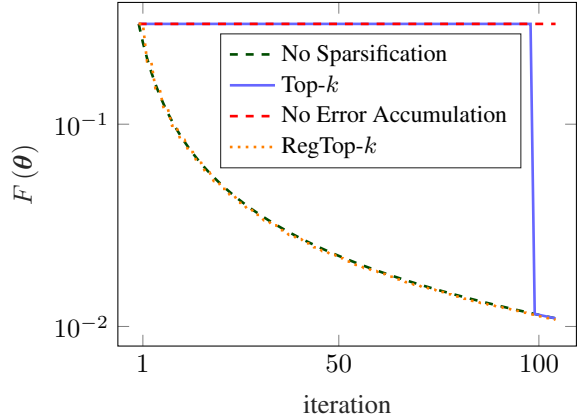}
	\end{center}
	\caption{Example of large learning rate scaling in \tpK. \textit{Since the largest local entries cancel out after aggregation at the server, \tpK makes no progress over many iterations.}}
\label{fig:toy}
\end{figure}

Fig.~\ref{fig:toy} shows the training loss against iterations for learning rate $\eta = 0.9$ and $\btheta^{0} = \dbc{0, 1}$ using \textsc{Top-}$1$ and \textsc{RegTop-}$1$ sparsification, as well as centralized training. We observe that \textsc{Top-}$1$ is not able to reduce the empirical risk even after $100$ iterations. This behavior comes from the heterogeneity of local gradients: at $\btheta^{0}$, the gradients are $\bg_1 = 0.736 \dbc{- 100,  1}$ and $\bg_2 = 0.736 \dbc{ 100,  1}$. \textsc{Top-}$1$ selects the largest entries, i.e., the first entries. However, despite their significantly larger amplitudes, the first entries do not contribute in the training, as they cancel out after averaging. With \textsc{Top-}$1$, the aggregated sparsified gradient remains zero, i.e., $0.736 \dbc{- 100,  0}+ 0.736\dbc{ 100,  0}$, and hence the distributed gradient descent remains at $\btheta^0$. The sparsified gradient descent will start to move from the initial point only after a large number of iterations, when the accumulated sparsification error at the second entry starts to surpass the first entry in amplitude, namely after the aggregated error at the second entry exceeds $100$. As the figure shows, the proposed \rgtpK scheme tracks ideal training without sparsification.

In this toy example, \tpK sparsification only delays the convergence. Nevertheless, one can observe that learning rate scaling potentially hinders the convergence in some other settings, e.g., leading to large optimality gap. To see this issue, consider a setting in which the loss of worker $n$ is $\tilde{F}_n \brc{\btheta} = F_n \brc{\btheta} + G\brc{\theta_2}$, for some $G\brc{\theta_2}$ whose derivative at $\theta_2=1$ is $1$. Starting from $\btheta^{0} = \dbc{0, 1}$, \textsc{Top-}$1$ aggregates zero gradients in the first $50$ iterations. The distributed gradient descent moves out of $\btheta^{0}$ only at $t = 51$ when the accumulation error at both workers read $\beps_n^{t} = \dbc{0,100}$. After moving out, the workers send their sparsified accumulated gradients $\hat{\bg}_n^{t} = \dbc{0,100}$, which lead to ${\bg}^t = \dbc{0,100}$. Comparing ${\bg}^t $ with non-sparsified aggregation in the first iteration, we see that \textsc{Top-}$1$ scales the learning rate with factor $50$. Depending on $G\brc{\theta_2}$, this can substantially deteriorate the convergence. 

\subsection{Contributions}
In this work, we develop a Bayesian framework for gradient sparsification. Unlike earlier studies, we focus on the learning rate scaling of \tpK and propose a regularization technique to control this property. To the best of our knowledge, this problem has not been considered in earlier studies. The core idea of our proposed scheme is well understood from the toy example presented in Section~\ref{sec:example}: it is readily observed that the better choice for sparsification is to send the gradient entries that constructively contribute aggregation. This choice can be made, if \textit{the workers infer the contribution of their gradient entries in the aggregated gradient}. Invoking the Bayesian framework, we propose a stochastic scheme to estimate this contribution from the previous globally-known aggregations. In summary, our main contributions are as follows:
\begin{itemize}
\item We formulate gradient sparsification as an inverse probability, i.e., statistical inference, problem. Invoking this formulation, we derive the Bayesian optimal sparsification mask via the \ac{map} estimator. The computation of this mask requires a prior belief on local gradients and characterization of the likelihood.  
\item To specify the prior belief on local gradients, we interpret the \tpK algorithm as a mismatched form of the optimal Bayesian sparsification algorithm. Invoking this interpretation, the prior distribution of gradient entries is considered to be proportional to the magnitude of accumulated gradient entries. We call this prior distribution the \tpK prior belief.
We then characterize the likelihood by describing the next aggregated model in terms of the former aggregations through a forward probability problem, i.e., a generative statistical model, with additive \textit{innovation}. We finally invoke large deviation arguments to derive the likelihood term when the model is large. 
\item Using the \tpK prior along with the asymptotic likelihood, we propose the \rgtpK algorithm whose sparsification mask selects the entries with maximum posterior probability of being within the top $k$ entries of the global gradient. We show that this sparsification mask can be observed as a regularized version of the \tpK mask, where the regularization can control the scaling of the learning rate. This observation further clarifies the reason behind appellation. %
\item We validate the proposed sparsification scheme through numerical experiments. The results on linear regression show that unlike \tpK that remains at a fixed distance from the global minimum, \rgtpK converges linearly to the optimal point.  The results for training ResNet-18 on CIFAR-10 with $0.1\%$ sparsification further depict up to $8\%$ higher classification accuracy as compared with \tpK. We further use \rgtpK to fine-tune SqueezNet \cite{iandola2016squeezenet}, ShuffleNetV2 \cite{ma2018shufflenet}, MobileNetV2 \cite{sandler2018mobilenetv2}, EfficientNet \cite{tan2019efficientnet}, and ResNet-152 on the ImageNette dataset \cite{imagenette} and compare it against \tpK sparsification. The numerical results demonstrate the \textit{statistical significance} of \rgtpK over \tpK for all sparsity levels.
\end{itemize}

\subsection{Further Related Work}
\label{sec:2}
Several lines of work have extended distributed \ac{sgd} with \tpK sparsification. The study in \cite{chen2018adacomp} proposes an adaptive sparsification technique based on \tpK aiming to enhance the computational complexity of distributed learning in large networks. The authors of \cite{lin2018deep} have developed the \textit{deep gradient compression scheme} that incorporates the ideas of momentum correction into \tpK. Online adaptation of \tpK with the goal of having minimum training time is further discussed in \cite{han2020adaptive}. In \cite{wang2018atomo}, the \tpK extension \textit{Atomo} is introduced, which sparsifies the gradients in an arbitrary atomic decomposition space, e.g., singular value and Fourier decomposition. 
Layer-wise gradient sparsification for \acp{dnn} is further investigated in \cite{zhang2022mipd}. 
The study in \cite{zhou2021efficient} proposes a new technique to speed up back propagation in sparsified distributed \ac{sgd}. In \cite{chen2020scalecom}, the authors propose a scalable sparsified gradient compression that uses the similarity of local gradients to enhance the scalability of \tpK. It is worth mentioning that the \textit{above lines of work are fundamentally different from our study} in the sense that they aim to adapt classical \tpK to a wider range of distributed settings, e.g., adapting it to other optimizers rather than basic \ac{sgd}. Unlike these lines of work, our study aims to develop a \textit{novel sparsification scheme} that controls learning rate scaling. 
%

Developing sparsification algorithms based on model statistics has been recently investigated in \cite{sahu2021rethinking} and \cite{m2021efficient}. In \cite{sahu2021rethinking}, the authors revisit \tpK sparsification and give an alternative interpretation for it as the communication-optimal sparsifier for per-iteration communication budget. This notion of optimality is then extended to the entire training leading to a new \tpK based sparsification scheme, referred to as hard-threshold sparsifier. The study in \cite{m2021efficient} moreover proposes a statistical approach for gradient sparsification by treating local gradients as random variables distributed with empirically-validated sparsity-inducing distributions. Invoking this stochastic model, the authors then propose a threshold-based sparsification technique with lower computation overhead. It is worth noting that the aim of these lines of work is to address practical challenges of \tpK, e.g., \cite{sahu2021rethinking} focuses on communication budget and \cite{m2021efficient} considers the computation overhead. This is fundamentally different from the goal of this study which tries to develop an alternative sparsification algorithm that uses model statistics to control the learning rate scaling caused by error accumulation. From this perspective, i.e., scaling control, these earlier studies behave identical to \tpK.
To the best our knowledge, this work is the first study that takes this issue into account and develops an \textit{alternative} sparsification algorithm based on error accumulation that \textit{controls} the impact of learning rate scaling.

\subsection{Notation}
Vectors are shown in bold, e.g., $\bxx$. The $j$-th entry of vector $\bxx$ is represented as $x\id{j}{}$. Entry-wise multiplication and division are denoted by $\odot$ and $\oslash$, respectively. The inner product of $\bxx$ and $\byy$ is represented by $\inner{\bxx;\byy}$. The $\ell_p$-norm of $\bxx$ is denoted by $\norm{\bxx}_p$. For an integer $N$, the set $\set{1,\ldots,N}$ is abbreviated as $\dbc{N}$. We denote the cardinality of set $\setS$ by $\abs{\setS}$.

\section{Preliminaries}
We consider the distributed setting described in Section~\ref{sec:intro}: let $\setD_n = \set{ \bxx_{n,i} \; \text{ for } \; i \in \dbc{D_n}  }$ denote the training mini-batch at worker $n\in\dbc{N}$ with $D_n = \abs{\setD_n}$ being the batch-size. Assume that the data-points in $\setD_n$ are sampled \ac{iid} from a distribution $p\brc{\bxx}$.  

\subsection{Gradient-Based Distributed Training}
The distributed training in this setting is formulated as
\begin{align}
\min_{\btheta\in \setR^J}  \dfrac{1}{\displaystyle\sum_{n=1}^N D_n} \sum_{n=1}^N D_n F_n(\btheta) \label{eq:main}
\end{align}
where $F_n(\btheta)$ is the empirical loss computed by worker $n$ over its (potentially randomly-samples) batch of data, i.e.,
\begin{align}
F_n\brc{\btheta} = \frac{1}{D_n} \sum_{i=1}^{D_n} f(\btheta\vert \bxx_{n,i})
\end{align}
for a loss function $f\brc{\btheta\vert \bxx}$ parameterized with model parameters $\btheta\in\setR^J$. The ultimate goal is to solve the optimization \eqref{eq:main} in a distributed fashion with minimal communication overhead. The server employs a general gradient-based optimizer: in iteration $t$, it computes a global gradient estimator from its observations on the local estimators $\bg_n^{t}$ for $n\in\dbc{N}$ received through the network.\footnote{In the ideal case, with perfect knowledge on the local estimators, the server computes its estimator as in \eqref{eq:glob} for some weights $\omega_n$. However, as explained below, it usually has only a compressed version of the local estimators.}  
It then uses the gradient estimator $\bg^t$ to update the model, e.g., in \ac{sgd} it applies $\btheta^{t+1} = \btheta^t - \eta^t \bg^{t}$. %

\subsection{\tpK: Classical Sparsification Scheme}
For communication efficiency, the workers invoke gradient sparsification. This means that every worker sends a \textit{compressed} form of its local gradient that is sparser. The classical algorithm for sparsification is \tpK presented in Algorithm~\ref{alg:topK}. In this algorithm, $\baa_n^t$ is the \textit{accumulated gradient} in iteration $t$, which is computed by adding the local gradient $\bg_n^t$ to the \textit{sparsification error} $\beps_n^{t}$ computed at the end  of iteration $t-1$; see line 3. The \textit{sparsification mask} of worker $n$, denoted by $\bss_n^t\in\set{0,1}^J$, is then computed by selecting the $k$ non-zero entries of $\baa_n^t$ with largest magnitudes; see line 4. We denote this operation by the \textit{top $k$ selector} $\Top{k}{\cdot}$, which is defined as follows: let $\abs{x_{i_1}} \geq \ldots  \abs{x_{i_J}}$ be the magnitude sorting for $\bxx\in\setR^J$; then, the $i$-th entry of $\Top{k}{\bxx}\in\set{0,1}^J$ is 
\begin{align}
\Top{k}{\bxx}_{\dbc{i}} = \begin{cases}
	1 &\text{if } \; i \in \set{i_1, \ldots, i_k}\\
	0 &\text{elsewhere}
\end{cases}.
\end{align}

\begin{algorithm}[tb]
\caption{\tpK Sparsification at Worker $n$}
\label{alg:topK}
\textbf{Initialization}: Set $\beps_n^0 = \set{0}^J$
\begin{algorithmic}[1] 
	\FOR{$t\geq 0$}
	\STATE Determine local gradient $\bg_n^t$ at global model $\btheta^t$
	\STATE Determine \textit{accumulated gradient} as $\baa_n^t = \beps_n^{t} + \bg_n^t$
	\STATE Find the sparsification mask as $\bss_n^t = \Top{k}{\baa_n^t}$
	\STATE Set the sparsified gradient to $\hat{\bg}_n^t =\bss_n^t \odot \baa_n^t$
	\STATE Send the non-zero entries of  $\hat{\bg}_n^t $ along with their indices to the server
	\STATE Update the \textit{sparsification error} as $\beps_n^{t+1} = \baa_n^{t} - \hat{\bg}_n^t$
	\ENDFOR
\end{algorithmic}
\end{algorithm}

Using the sparsification mask, worker $n$ sparsifies its gradient as $\hat{\bg}_n^t =\bss_n^t \odot \baa_n^t$ and send it to the server. The server then estimates the gradient of the empirical loss as 
\begin{align}
	{\bg}^t = \sum_{n=1}^N \omega_n \hat{\bg}_n^t
\end{align}
and updates the model as $\btheta^{t+1} = \btheta^t - \eta^t {\bg}^{t}$. Note that the communication reduction by gradient sparsification is achieved at the expense of an extra index transmission per symbol. However, the index can be losslessly represented by $\log J$ bits, so its communication load can be neglected. The \textit{sparsification factor} of \tpK, i.e., the factor by which the communication load decreases, is hence approximately $S \approx k/J$. In practice, $S$ is in order of $0.01$ \cite{lin2018deep,sahu2021rethinking}.

\section{Bayesian Gradient Sparsification}
\label{sec:Bayes}
The key contribution of this study is the development a Bayesian framework for gradient sparsification, which leads to derivation of \rgtpK presented in Section~\ref{sec:RegTopK}. In this section, we present this framework in two major steps:
\begin{enumerate}
\item We first formulate the sparsification task as an inference problem whose Bayesian optimal solution is given by a  \ac{map} estimator. We then use the analogy to classical \tpK to specify the prior belief of this problem.
\item We  use the large-deviation arguments to characterize the likelihood in the asymptotic regime. Using this asymptotic characterization, we derive a generic expression for the sparsification mask of the Bayesian optimal sparsifier. 
\end{enumerate}

In the sequel, we go through these steps in greater detail. The  \rgtpK algorithm, which is the direct conclusion of this Bayesian framework, is then presented in the next section.

\subsection{Statistical Global \tpK}
Let us start with an imaginary scenario in which a genie provides the workers information about the aggregated gradients of the other workers. For entry $j \in \dbc{J}$, suppose worker $n$ knows in advance 
$a\id{j}{}^t$, which is the $j$-th aggregated symbol when the workers apply no sparsification, i.e., 
\begin{align}
	a\id{j}{}^t = \sum_{n=1}^N  \omega_n a\id{j}{n}^t.
\end{align}
In such case, worker $n$ decides to transmit $a\id{j}{n}^t$, only if $a\id{j}{}^t$ is within the top $k$ gradient entries. In other words, 
they apply \tpK directly on the averaged gradient. We refer to this idealized approach as \textit{global \tpK}. 

Global \tpK is infeasible, since the workers do not know $a\id{j}{}^t$. Nevertheless, they have partial\footnote{Note that the workers sparsify their gradients. This means that some entries may be never sent to the server.} access to $a\id{j}{}^t$ for $\ell < {t}$. This can be used to estimate $a\id{j}{}^t$, i.e., the workers use information collected through time to apply global \tpK statistically. In a Bayesian framework, we can formulate a \textit{statistical global \tpK} via the principle \ac{map} problem defined below. 

\begin{definition}[Principle MAP Problem]\label{def:MAP}
Let $\setT_k^t$ denote the set of $k$ largest entries of $\abs{\baa^t}$ that is unknown to all the workers. Worker $n$ determines its posterior probabilities
\begin{align}
	P\id{j}{n}^t = \Pr\set{j \in \setT_k^t \left\vert \baa_n^t, \set{\baa_n^\ell, \bg^\ell: \ell < {t} } \right.}
\end{align}
for  $j\in\dbc{J}$, where $\bg^\ell$ is the aggregated gradient in iteration $\ell < t$ that is known to all workers. Worker $n$ then selects the $k$ entries with the largest posteriors.
\end{definition}

To solve the principle \ac{map} problem, we expand the posterior probability as 
\begin{align}
	P\id{j}{n}^t &= 
	\mal\id{j}{n}^t
	\Pr\set{j \in \setT_k^t \left\vert \baa_n^t  \right.} \label{eq:bayes}
\end{align}
where $\Pr\set{j \in \setT_k^t \left\vert \baa_n^t  \right.}$ describes the \textit{prior belief}, i.e., the prior assumption on the probability of $j \in \setT_k^t$ based on the local \textit{accumulated} gradient of worker $n$, and $\mal\id{j}{n}^t$ denotes the \textit{likelihood} that is given by
\begin{align}
\mal\id{j}{n}^t \propto p\brc{ \left. { \baa_n^\ell, \bg^\ell: \ell< {t} } \right\vert  j \in \setT_k^t ,  \baa_n^t}.
\end{align}
Here, we use notation $p\brc{\cdot}$ to refer to the \ac{pdf}. 
While the likelihood is determined by the forward probability problem, the prior belief is specified by a postulated model on the distribution of global \tpK entries.

The standard \tpK algorithm considers only the local accumulated gradients $\baa_n^t$, which suggests that its prior belief is proportional to the magnitude of the local accumulated gradient. This is formalized as follows.
\begin{definition}[\tpK Prior Belief]\label{def:Prior}
Given the accumulated gradient $\baa_n^t$, \tpK applies the prior belief that the probability of entry $j$ being among the top $k$ entries of $\bg^t$ is proportional to $\abs{a\id{j}{n}^t}$, i.e., 
\begin{align}
	\Pr\set{j \in \setT_k^t \left\vert \baa_n^t  \right.} = \frac{\abs{a\id{j}{n}^t}}{\norm{\baa_n^t}_1}. \label{eq:prior}
\end{align}
\end{definition}
\noindent With this prior belief, the \ac{map} selector reduces to
\begin{align}
\argmaxk_j P\id{j}{n}^t 
&= \argmaxk_j
\mal\id{j}{n}^t
\abs{a\id{j}{n}^t} \label{eq:map-general},
\end{align}
where $\argmaxk_j x_j$ returns the $k$ largest entries of the sequence $\set{x_j: j \in \dbc{J}}$ for $J\geq k$.

Comparing the Bayesian sparsifier in \eqref{eq:map-general} with \tpK, one can interpret \textit{\tpK as a \ac{map} selector whose likelihood $\mal\id{j}{n}^t$ is \textit{uniform}}. This is in general a \textit{mismatched} algorithm: \tpK simply ignores the time series collected in previous communication rounds and infer the dominant entries solely based on local gradients. Considering this alternative formulation, our goal is to improve \tpK by designing a more appropriate likelihood. 

\subsection{Characterizing the Likelihood} 
\label{sec:like}
It is not straightforward to determine the exact expression for likelihood, since the statistical model of its forward probability problem is not completely known. We hence follow an alternative approach, in which we characterize the likelihood asymptotically under a set of simplifying assumptions and utilize it to approximate the likelihood term.  
Let us start the derivations by presenting an alternative expression for $P\id{j}{n}^t $.
\begin{proposition}
	\label{prop:1}
The posterior probability $P\id{j}{n}^t $ is alternatively computed as 
\begin{align*}
	P\id{j}{n}^t  
	= \int_{\setF_j^k }
	q_n({\baa^t}) \dif \baa^t,
\end{align*}
where $\setF_j^k \subset \setR^J$ is the set of points in $\setR^J$ whose $j$-th entries are among their largest $k$ entries and $q_n({\baa^t})$ is 
\begin{align}
	q_n({\baa^t}) = 
	p \brc{ \baa^t \left\vert \baa_n^t,  \set{\baa_n^\ell, \bg^\ell: \ell < {t} } \right.}. \label{eq:q}
\end{align}
\end{proposition}
\begin{proof}
We start by marginalizing the posterior probability over the \textit{unknown} global accumulated gradient $\baa^t$ as\footnote{The derivation follows a similar idea as the one used in Kalman filtering. This is intuitive considering the connection between the two problems.}
\begin{subequations}
	\begin{align}
		P\id{j}{n}^t  &= \Pr\set{j \in \setT_k^t \left\vert \baa_n^t,  \set{\baa_n^\ell, \bg^\ell: \ell < {t} } \right.} \label{eq:newMAP}\\
		&= \int  Q\id{j}{n}^t \brc{\baa^t}
		q_n({\baa^t}) \dif \baa^t ,
	\end{align}
\end{subequations}
with $q_n({\baa^t})$ being defined in \eqref{eq:q} and
\begin{align}
	Q\id{j}{n}^t \brc{\baa^t} = \Pr\set{j \in \setT_k^t \left\vert  \baa_n^t, \baa^t,  \set{\baa_n^\ell, \bg^\ell: \ell< {t} } \right.}.
\end{align}
As $\setT_k^t$ contains global top-$k$ entries, it is deterministically computed from $\baa^t$.  Thus, $\baa_n^t,  \set{\baa_n^\ell, \bg^\ell :\ell < t} \rightarrow \baa^t \rightarrow \setT_k^t$ form a Markov chain, and
\begin{align}
	Q\id{j}{n}^t \brc{\baa^t}  = \mathds{1} \set{ \baa^t \in  \setF_j^k }, 
\end{align}
where $\mathds{1} \set{\cdot}$ denotes the indicator function, and $\setF_j^k \subset \setR^J$ is defined as in the proposition. 
For instance, in the simple case of $J = 2$ and $k=1$, we have two such sets, namely
\begin{align*}
	\setF_1^k &=\set{\bx \in \setR^2: \abs{x_1} > \abs{x_2}}\\
	\setF_2^k &=\set{\bx \in \setR^2: \abs{x_2} > \abs{x_1}}.
\end{align*}
A visualization for this simple example is
is given in Fig.~\ref{fig:P_region}.

\begin{figure}
	\begin{center}
			\begin{tikzpicture}
	\begin{axis}[
		width=2.3in, 
		height=2.3in, 
		axis lines = middle,
		xlabel = {$x_1$},
		ylabel = {$x_2$},
		xmin=-6, xmax=7,
		ymin=-6, ymax=7,
		xtick={8}  ,
		xticklabels={{}},
		ytick={8}  ,
		yticklabels={{}},
		]
		
		\addplot [name path = A,
		dashed,
		domain = -5:5,
		samples = 1000] {x};
		
		\addplot [name path = t,
		domain = -6:6,
		samples = 1000] {0}
		node [pos=0.8, above] {$\setF^k_1$};
		
		\addplot [name path = t2,
		draw=none,
		domain = -1:0,
		samples = 1000] {3}
		node [right] {$\setF^k_2$};

		\addplot [name path = B,
		dashed,
		domain = -5:5] {-x};
		
		\addplot [name path = C,
		draw=none,
		domain = -5:5,
		samples = 1000] {4.5};
		
		\addplot [name path = D,
		draw=none,
		domain = -5:5,
		samples = 1000] {-4.5};
		
		\addplot [red!20] fill between [of = B and A, soft clip={domain=-4.5:4.5}];
		\addplot [blue!20] fill between [of = C and A, soft clip={domain=0:4.5}];
		\addplot [blue!20] fill between [of = C and B, soft clip={domain=-4.5:0}];
		\addplot [blue!20] fill between [of = D and B, soft clip={domain=0:4.5}];
		\addplot [blue!20] fill between [of = D and A, soft clip={domain=-4.5:0}];
		
	\end{axis}
\end{tikzpicture}

%
%
%
%
		\end{center}
	\caption{The feasible regions $\setF_{1}^k$ and $\setF_{2}^k$ for $J=2$ and $k=1$.}
	\label{fig:P_region}
\end{figure}
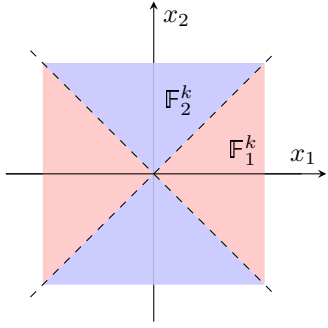 

Substituting in \eqref{eq:newMAP} we can write
\begin{align}
	P\id{j}{n}^t  
	&= \int   \mathds{1} \set{ \baa^t \in  \setF_j^k }
	q_n({\baa^t})  \dif \baa^t
	= \int_{\setF_j^k }
	q_n({\baa^t}) \dif \baa^t
\end{align}
which concludes the alternative form.
\end{proof}



Comparing this alternative form of $P\id{j}{n}^t$ to the one derived via the Bayes rule, we can write 
\begin{align}
\mal\id{j}{n}^t \propto \frac{1 }{\abs{a\id{j}{n}^t}} \int_{\setF_j^k }
q_n({\baa^t}) \dif \baa^t. \label{eq:Likelihood}
\end{align}
The likelihood calculation is hence reduced to characterization of the conditional distribution $q_n\brc{\baa^t}$, which 
 requires an explicit statistical model for time dependency of training gradients. Such information is however neither tractably acquired nor universal. To develop a universal algorithmic framework, we describe $q_n({\baa^t})$ under some simple model: we consider a stochastic model whose inputs are $\baa_n^t$ and  $\set{\baa_n^\ell, \bg^\ell: \ell < {t} }$, and whose output (observation) is $\baa^t$. 
 
 To describe this model, we start by writing  
\begin{align}
\baa^t = \omega_n \baa_n^t + \bzz_n^t \label{eq:1}
\end{align}
where $\bzz_n^t$ denotes the average of all local gradients with worker $n$ being excluded. We now assume that the time evolution of $\bzz_n^t$ is described via an additive model, i.e.,  %
\begin{align}
\bzz_n^t =  \bzz_n^{t-1} + \bxi_n^{t}
\end{align}
for some \textit{innovation} $\bxi_n^{t}$, which we treat as a random variable. Noting that the innovation describes the difference between two consecutive local gradients, it is realistic to assume that $\bxi_n^{t}$ is heavily distributed within a small vicinity of zero. 

We next divide the index set $\dbc{J}$ into two subsets, namely $\setS_n^{t-1}$ that denotes the support of the sparsification mask $\bss_n^{t-1}$ computed by worker $n$ in the previous iteration,~and~its~complement, i.e., $\dbc{J}-\setS_n^{t-1}$. Let us first focus on $j \in \setS_n^{t-1}$. For this set, worker $n$ has already received the aggregated gradient in the last iteration, i.e., $a\id{j}{}^t = \rmg \id{j}{}^t$. We can hence write 
\begin{align}
z\id{j}{n}^{t-1}  
&= \rmg\id{j}{}^{t-1} - \omega_n a\id{j}{n}^{t-1}
= \omega_n a\id{j}{n}^{t} \Delta\id{j}{n}^{t} 
\end{align}
where we define 
\begin{align}\label{eq:Delta_def}
\Delta\id{j}{n}^{t} = \frac{\rmg\id{j}{}^{t-1} - \omega_n a\id{j}{n}^{t-1}}{\omega_n a\id{j}{n}^{t}},
\end{align}
and refer to it as the \textit{posterior distortion}. By substituting into \eqref{eq:1}, we have $a\id{j}{}^t = \bar{\rmg}\id{j}{n}^t + \xi\id{j}{n}^{t}$, where $ \bar{\rmg}\id{j}{n}^t $ is given by
\begin{align}
\bar{\rmg}\id{j}{n}^t = \omega_n a\id{j}{n}^t \brc{1+\Delta\id{j}{n}^{t} } .
\end{align}
Considering this postulated model, $a\id{j}{}^t$ for $j \in \setS_n^{t-1}$ is conditionally distributed with the innovation distribution shifted by $\bar{\rmg}\id{j}{n}^t$. For sparsified entries, i.e., $j \notin \setS_n^{t-1}$, $z\id{j}{}^{t-1}$ is missing. We hence write\footnote{Considering this model for all entries, we recover the \tpK algorithm.}
\begin{align}
	a\id{j}{}^t = \omega_n a\id{j}{n}^t + \xi\id{j}{n}^{t} .
\end{align}

Given an innovation distribution, one can determine the likelihood from \eqref{eq:Likelihood} using the time evolution model postulated for local gradients. Nevertheless, even for simple innovation distributions, determining an analytic expression for the likelihood, that can be efficiently computed for iteration-level sparsification, is not tractable. We can however invoke large deviations arguments \cite{dembo2009large} and the method of types \cite{csiszar1998method} to approximate the likelihood for some fast-decaying innovation distributions. This approximation is presented in the following result whose derivation is given in the next sub-section.

\begin{result}
	\label{prop:2}
Let entries of $\bxi_{n}^{t}$ be independent and $p_j \brc{\xi}$ denote the distribution of the $j$-th entry in $\bxi_{n}^{t}$, i.e., $ \xi\id{j}{n}^{t}$. Assume that $p_j \brc{\xi}$ is symmetric and zero-mean for $j\in \dbc{J}$, and that given a small $\delta$, there exists a small $\varepsilon$, such that 
\begin{align}
	\int_{-\varepsilon \abs{a\id{j}{n}^t}}^{\varepsilon \abs{a\id{j}{n}^t} } p_j \brc{\xi} \dif \xi \geq 1 - \delta.
\end{align}
Then, as the model size grows large, i.e.,  $J\to \infty$, under some heuristics, we have
	\begin{align}
	\mal\id{j}{n}^t \propto
		\begin{cases}
			u_\mu ({\abs{ 1 +   \Delta\id{j}{n}^{t}  } }) &j \in \setS_n^{t-1}\\
			C &j \notin \setS_n^{t-1}
		\end{cases},
	\end{align}
	for some constant $C$ and some function $u_\mu\brc\cdot$ that is a cumulative distribution parameterized by $\mu$. 
\end{result}

The above results suggests that similar \tpK for \textit{unselected entries} $j \notin \setS_n^{t-1}$, the likelihood is fixed in $j$. However, for those selected in the last iteration, i.e., $j \in \setS_n^{t-1}$, the likelihood is of the form
\begin{align}
	\mal\id{j}{n}^t = u_\mu ({\abs{ 1 +   \Delta\id{j}{n}^{t}  } }), 
\end{align}
which varies against $j$.

\subsection{Deriving Result~\ref{prop:2}}
Recall that $\bzz_n^t$ represents the aggregated accumulated gradient of all workers excluding worker $n$, i.e., 
\begin{align}
	\bzz_n^t = \sum_{m\neq n} \omega_m \baa_m^t.
\end{align}
The aggregated accumulated gradient in iteration $t$, which is unknown to the workers, is hence given by 
\begin{align}
	\baa^t &= \omega_n \baa_n^t + \bzz_n^t =  \omega_n \baa_n^t + \bzz_n^{t-1} + \bxi_n^{t}.
\end{align}
Note that $\bzz_n^{t-1}$ is known at those entries that have been selected by the sparsification mask in iteration $t-1$, i.e., $j \in \setS_n^{t-1}$ with $\setS_n^{t-1}$ being the support of the last sparsification mask $\bss_n^{t-1}$.

Proposition~\ref{prop:1} gives the posterior probability of the $j$-th entry in terms of the conditional distribution $q\brc{\baa^t}$ and the feasible set ${\setF_j^k }$. %
Recall that the feasible set ${\setF_j^k }$ is the set of points in $\setR^J$ in which the $j$-th element is among the top $k$ elements.

To characterize the conditional distribution $q\brc{\baa^t}$, we first utilize the marginalization rule to write
\begin{subequations}
	\begin{align}
		q\brc{\baa^t} &= p\brc{ \baa^t \left\vert \baa_n^t,  \set{\baa_n^\ell, \bg^\ell: \ell < {t} } \right.}\\
		&=  \int   q\brc{ \baa^t , \bzz_n^{t-1} }  p \brc{ \bzz_n^{t-1} \left\vert \baa_n^t,  \set{\baa_n^\ell, \bg^\ell: \ell < {t} } \right.} \dif \bzz_n^{t-1}
	\end{align}
\end{subequations}
where we define $q\brc{ \baa^t , \bzz_n^{t-1} } $ as
\begin{align}
	q\brc{ \baa^t , \bzz_n^{t-1} } =  p\brc{ \baa^t \left\vert \bzz_n^{t-1}, \baa_n^t,  \set{\baa_n^\ell, \bg^\ell: \ell < {t} } \right.}.
\end{align}
Noting that $\set{\baa_n^\ell, \bg^\ell: \ell < {t} } \to \brc{\bzz_n^{t-1}, \baa_n^t} \to \baa^t $ form a Markov chain, we can write
\begin{subequations}
	\begin{align}
		q\brc{ \baa^t , \bzz_n^{t-1} }  &= 
		p\brc{ \baa^t \left\vert \bzz_n^{t-1}, \baa_n^t  \right.}\\
		&= p_{\bxi_n}^t \brc{ \baa^t - \omega_n \baa_n^t - \bzz_n^{t-1} },
	\end{align}
\end{subequations}
where $p_{\bxi_n}^t \brc{\cdot}$ denotes the \ac{pdf} of $\bxi_n^t$. Substituting in Proposition~\ref{prop:1}, the posterior probability is given by
\begin{align}
	P\id{i}{n}^t  
	=   \int   I_i\brc{\bzz_n^{t-1}}   q_t\brc{\bzz_n^{t-1} } \dif \bzz_n^{t-1},
\end{align}
where $I_i\brc{\bzz_n^{t-1}} $ is given by
\begin{align}
	I_i\brc{\bzz_n^{t-1}}  
	&=	\int_{\setF_i^k }  p_{\bxi_n}^t \brc{ \baa^t - \omega_n \baa_n^t - \bzz_n^{t-1} } \dif \baa^t,
\end{align}
and $q_t\brc{\bzz_n^{t-1} }$ is defined as
\begin{align}
	q_t\brc{\bzz_n^{t-1} }
	=    p \brc{ \bzz_n^{t-1} \left\vert \baa_n^t,  \set{\baa_n^\ell, \bg^\ell: \ell < {t} } \right.} \dif \bzz_n^{t-1}.
\end{align}

In the Appendix~\ref{append:B}, we invoke large deviations arguments and the method of types to approximate $I_i\brc{\bzz_n^{t-1}}$ for large~$J$. Our analysis relies on some heuristic assumptions, including the assumption that the innovation variance, i.e., the variance of distribution $p_i\brc{\cdot}$, scales with the accumulated gradient. Under these assumptions, we show that for large $J$
\begin{align}
	I_i\brc{\bzz_n^{t-1}} \approx K \abs{a\id{i}{n}^t} u_{\mu} \brc{\abs{1+\Delta\id{i}{n}^t}},
\end{align}
for some constant $K$ and $\mu$. The term $u_\mu \brc{\cdot}$ in this expression is the cumulative distribution of an scaled and shifted form of the innovation variable, i.e., entries of $\bxi_n^t$, with the parameter $\mu$ specifying the distribution, e.g., the variance. Note that $u_\mu \brc{x}$ is increasing in $\abs{x}$ whose minimum occurs at $x=0$.

%
%

%

We recall that $\bzz_n^{t-1}$ is not entirely known, as worker $n$ can only compute $z\id{i}{n}^{t-1}$ for those entries that have been selected in round $t-1$, i.e., $i \in \setS^{t-1}$. We hence postulate that  
\begin{align}
	q_t\brc{z\id{i}{n}^{t-1} }
	= \begin{cases}
		\mathds{1} \set{
			z\id{i}{n}^{t-1} = \rmg\id{i}{}^{t-1} - \omega_n a\id{i}{n}^{t-1} 
		} &i \in \setS^{t-1}\\
		p_0 \brc{z\id{i}{n}^{t-1} }  &i \notin \setS^{t-1}
	\end{cases}
\end{align}
for some distribution $p_0 \brc{ \cdot}$. This postulated distribution can be interpreted as follows: for $i \in \setS^{t-1}$ the term $z\id{i}{n}^{t-1}$ is explicitly computed  from the model in Section~\ref{sec:like}. For those entries that are not sent, the worker considers a prior assumption on $z\id{i}{n}^{t-1}$. Noting that $z\id{i}{n}^{t-1}$ contributes in $I_i\brc{\bzz_n^{t-1}}$ through $\Delta\id{i}{n}^{t-1}$, we can conclude that for $i \in \setS^{t-1}$
\begin{align}
	P\id{i}{n}^t  \approx K \abs{a\id{i}{n}^t} u_{\mu} \brc{\abs{1+\Delta\id{i}{n}^t}},
\end{align}
while for $i \notin \setS^{t-1}$, we have
\begin{subequations}
	\begin{align}
	P\id{i}{n}^t  &\approx K \abs{a\id{i}{n}^t}
	\Ex{u_{\mu} \brc{\abs{1+\Delta\id{i}{n}^t}}}{\Delta\id{i}{n}^t}\\
	&= \hat{K} \abs{a\id{i}{n}^t},
\end{align}
\end{subequations}
for some $\hat{K}$ that is in general different from $K$. This implies
\begin{align}
	P\id{i}{n}^t  &\propto
	\begin{cases}
		\abs{a\id{i}{n}^t} u_\mu \brc{\abs{1+\Delta\id{j}{n}^t}} &i \in \setS^{t-1}\\
		C \abs{a\id{i}{n}^t}  &i \notin \setS^{t-1}
	\end{cases},
\end{align}
for some $C$. Using \eqref{eq:Likelihood}, we conclude Result~\ref{prop:2}.

\subsection{Discussions on Result~\ref{prop:2}}
Although Result~\ref{prop:2} is derived under some heuristics, it sheds some light on \tpK which is worth mentioning:
\begin{enumerate}
	\item Considering the Bayesian interpretation, \tpK assumes a prior mask distribution that is proportional to the magnitude of the accumulated gradient ${ \abs{a\id{i}{n}^t}}$. Result~\ref{prop:2} implies that this is equivalent to assuming that the innovation variance, i.e., variance of $\bxi_n^{t}$, scales with the accumulated gradient. This is intuitive, since with larger gradients the variation in the operating point increases.
	\item For \tpK sparsification, it is not required to restrict the order with which the prior grows with the magnitude of accumulated gradient, i.e., the exponent $y$ in ${ \abs{a\id{i}{n}^t}}^y$. In Definition~\ref{def:Prior}, we assumed that this order is 1. Nevertheless, we note that for any $ y \in (0,1]$ the Bayesian formulation with uniform likelihood leads to \tpK sparsification. This follows from the fact that with \textit{uniform} likelihood the sorting becomes independent of $y$. This is however not the case in the regularized version. We hence conjecture that by incorporating the hyper-parameter $y$ into \rgtpK, we can further improve the sparsification performance; see Remark~\ref{rmk:3} in Appendix for more detail.
	\item The derivation of \rgtpK elaborates the meaning of the uniform likelihood assumed by \tpK. As seen in the derivation, this means that \tpK ignores the residual terms $z\id{j}{n}^{t-1}$ for all entries. This is clearly sub-optimal and justifies our earlier intuitive discussion.
\end{enumerate}

\subsection{Bayesian Optimal Mask}
Invoking Result~\ref{prop:2}, we can now specify the sparsification mask for the \ac{map} sparsifier. The sparsification mask in this case is determined by applying \tpK on 
\begin{align}
	\hat{\baa}_n^t =  \baa_n^t \odot u_{\mu} \brc{ \abs{1+ \bdelta_n^t} }
\end{align}
for a cumulative distribution function $u_{\mu} \brc\cdot$. This can be interpreted as a form of regularization, in which the accumulated gradient entries are scaled with th asymptotic likelihood.  The effectiveness of the proposed mask is intuitively explained as follows: while for those gradient entries whose corresponding earlier aggregations are \textit{not} available, \tpK is efficient, for the other entries, earlier aggregations can be utilized to scale the gradients.  

We remark that the regularization scale computed by the \ac{map} sparsifier is proportional to 
\begin{align}
	r\id{j}{n}^{t} =  \left\vert { 1 +   \Delta\id{j}{n}^{t}  } \right\vert.
\end{align}
As $\Delta\id{j}{n}^{t} \to -1$, the regularization scale goes to its minimum value, i.e., $r\id{j}{n}^{t} \to 0$. Noting the cumulative distribution is increasing, this means that the scale in this case is minimized. This can be interpreted as follows: as the posterior distortion goes to $ -1$, worker $n$ infers that its $j$-th gradient entry will likely be canceled after aggregation, since it is canceled in the previous iteration, and the next aggregation is not expected to be considerably different from the one in the last iteration. The worker hence dampens this entry maximally to reduce its selection chance.\footnote{One can see this intuition explicitly by going through the motivational toy example in Section~\ref{sec:example}.}


\section{ \rgtpK Algorithm}
\label{sec:RegTopK}
\rgtpK is concluded from Result~\ref{prop:2} by specifying $u_\mu\brc\cdot$. As mentioned, 
$u_\mu\brc\cdot$ is determined by the innovation distribution  and $\mu$ is a parameter of this distribution. We hence treat it as a hyperparameter whose tuning is carried out through validation. Though Result~\ref{prop:2} suggests that $u_\mu\brc\cdot$ needs to describe a fast-decaying distribution, our numerical experiments demonstrate that using $\tanh$-type decaying is sufficient. This means that we postulate the innovation distribution to be
\begin{align*}
	p\brc{\xi} = \frac{1}{2\mu} \brc{1-\tanh^2 \frac{\xi}{\mu}}
\end{align*}
which decays slower than conventional choices, e.g., Gaussian. For this distribution, the cumulative function is given by
\begin{align}
	u_\mu \brc{x} = \frac{1}{2}\brc{ 1+ \tanh \frac{x}{\mu} }
\end{align}
and the standard deviation $\sigma$ is computed as $\sigma \approx 0.9 \mu$. Noting that the Bayesian sparsification in Definition~\ref{def:MAP} requires only relative magnitudes of the likelihoods, we can further drop the constants and set 
\begin{align}
	\mal\id{j}{n}^t  &\propto
	\begin{cases}
		\tanh \brc{\dfrac{\abs{1+\Delta\id{j}{n}^t}}{\mu}} &j \in \setS^{t-1}\\
		C &j \notin \setS^{t-1}
	\end{cases},
\end{align}
for a constant $C$ that can be treated as a hyperparameter.\footnote{Experiments show that setting $C = 1$ is effective. This corresponds to $u_{\mu}\brc{Q}$ for $Q\to \infty$, i.e., infinite posterior distortion, which is intuitive as in this case we have no information from the last iteration. } 
%
%
This concludes the \rgtpK algorithm. It is worth noting that the current choice of $u_\mu \brc{\cdot}$ is based on the postulated innovation model that is motivated by numerical investigations, and our derivation of \rgtpK is valid for other choices of $u_\mu \brc{\cdot}$. 

%
%

\rgtpK is given in Algorithm~\ref{alg:RegtopK}. The algorithm starts with applying standard \tpK in the initial communication round. Starting from $t=1$, worker $n$ after computing its accumulated gradient $\baa_n^t$ determines the {posterior distortion} $\bdelta_n^t$ for those entries that were sent in the previous round, i.e., $j$ for which $s\id{j}{n}^t =1$.\footnote{Recall that $s\id{j}{n}^t$ denotes the $j$-th entry of sparsification mask $\bss_n^t$.} This distortion is the key metric that is used by \rgtpK to regularize \tpK. 
Though shown in Section~\ref{sec:Bayes} through derivation, the role of posterior distortion can be further explained intuitively as follows: the posterior distortion determines the share of the other workers in the aggregated model relative to worker $n$. When $\Delta\id{j}{n}^t$ is close to $-1$, we can infer that the gradients of other workers aggregate to $- \omega_n a\id{j}{n}^t$ and will cancel the gradient of worker $n$.
%
This is better understood by considering two extreme cases:
\begin{itemize}
	\item[(1)] As $\mu \rightarrow 0$, the regularizer converges to $1$, and hence \rgtpK reduces to standard \tpK. \tpK can hence be seen as a special case of \rgtpK with no regularization; this further explains the appellation.
	\item[(2)] Another limiting case is a setting in which the $j$-th local gradient entries of all workers are large in amplitude but cancel each other out after aggregation.\footnote{Recall the toy example in Section~\ref{sec:example}.} In this case, after the initial aggregation, worker $n$ determines its posterior distortion as $\Delta\id{j}{n}^t = 0- a\id{j}{n}^{t-1}/ a\id{j}{n}^{t} = -1$. This leads to the $j$-th regularized accumulated gradient entry damped to zero and prevents its selection in the next iteration. This increases the frequency of selecting constructively-aggregating gradient entries with small amplitudes, avoiding large scaling of the learning rate.	
\end{itemize}


\begin{algorithm}[tb]
	\caption{\rgtpK Sparsification at Worker $n$}
	\label{alg:RegtopK}
	\textbf{Initialization}: Set $\beps_n^0 = \set{0}^J$, some $\mu > 0$, and a very large constant $Q$, i.e., $Q \rightarrow \infty$.
	
	\begin{algorithmic}[1] 
		\FOR{$t =  0$}
		\STATE Sparsify via \tpK
		\STATE Collect the aggregated model $\bg^0$ from the server
		\ENDFOR
		\FOR{$t\geq 1$}
		\STATE Determine local gradient $\bg_n^t$ at global model $\btheta^t$
		\STATE Determine \textit{accumulated gradient} as $\baa_n^t = \beps_n^{t} + \bg_n^t$
		\STATE Determine the \textit{posterior distortion} as
		\begin{align*}
			\hspace*{-2mm}\bdelta_n^t \hspace*{-.7mm}=\hspace*{-.7mm} \bss_n^{t-1}\odot\dbc{ \brc{\bg^{t-1} - \omega_n \baa_n^{t-1} } \oslash \omega_n \baa_n^{t} } \hspace*{-.7mm}+\hspace*{-.7mm} Q \brc{1-\bss_n^{t-1}}
		\end{align*}
		\STATE Find the sparsification mask as
		\begin{align*}
			\bss_n^t = \Top{k}{ \baa_n^t \odot \tanh \brc{ \dfrac{\abs{1+ \bdelta_n^t}}{\mu}} }
		\end{align*}
		\STATE Set the sparsified gradient to $\hat{\bg}_n^t =\bss_n^t \odot \baa_n^t$
		\STATE Send the non-zero entries of  $\hat{\bg}_n^t $ along with their indices to the server
		\STATE Update the \textit{sparsification error} as $\beps_n^{t+1} = \baa_n^{t} - \hat{\bg}_n^t$
		\ENDFOR
	\end{algorithmic}
\end{algorithm}

\begin{remark}
	From a computational perspective, \rgtpK and \tpK share the same order of complexity. Compared with \tpK, Algorithm~\ref{alg:RegtopK} introduces only minor additional overhead: the computation of the posterior distortion in line 8, and its element-wise multiplication (after activation) with the accumulated gradient prior to applying \tpK sparsification. These additional operations incur negligible cost and do not affect the overall complexity order. Remarkably, as demonstrated in the next section, despite having comparable computational complexity, \rgtpK significantly outperforms \tpK, particularly at high compression rates.
\end{remark}

\section{Numerical Investigation}
\label{sec:Num}
We validate \rgtpK through three sets of experiments: 
\begin{enumerate}
	\item We first consider the problem of linear regression, where we can explicitly calculate the global optimum, and thus characterize the optimality gap. 
	\item We compare \rgtpK against \tpK when used to train ResNet-18 on the CIFAR-10 dataset. 
	\item We use \rgtpK to fine-tune pretrained SqueezNet, ShuffleNetV2, MobileNetV2, EfficientNet, and ResNet-152 on ImageNette and compare the results against \tpK.
\end{enumerate}
Throughout the numerical experiments, we compare \rgtpK against two benchmarks: ($i$) \tpK, and ($ii$) the distributed \ac{sgd} without sparsification. In is worth reminding that as mentioned in the introduction, the existing improvements to \tpK extend it in orthogonal respects. For instance, \cite{sahu2021rethinking} extends the notion of optimality, and \cite{m2021efficient} proposes approaches to speed up the training process. Hence, \tpK also represents the behavior of its existing extensions. In fact, when it comes to the hindrance caused by large learning rate scaling, these approaches perform identical to \tpK.

\subsection{Linear Regression Problem}
\label{sec:linReg}
We start with the linear regression problem. Our interest comes from the following fact: we can track its optimal solution and hence compute the \textit{optimality gap, i.e., the distance between the model and global optimum}, which is~a~strong~notion to evaluate the convergence. This section presents key numerical findings. Further illustrative examples are given in Appendix~\ref{append:A}. 

We consider a distributed setting with $N=20$ workers that solve a distributed linear regression problem via distributed \ac{sgd}. Each worker has $D_n = 500$ labeled data-points, i.e.,  
\begin{align}
\setD_n = 	\set{\brc{\bxx_{n,i} , y_{n,i}} \in \setR^{J} \times \setR:  i \in \dbc{D_n}},
\end{align}
for $J=100$. We represent the local dataset $\setD_n$ compactly by matrix $\mX_n \in \setR^{D_n \times J}$ whose $i$-th row represents the $i$-th data-point $\bxx_{n,i}$, and vector $\by_n \in \setR^{D_n}$ whose $i$-th entry denotes the $i$-th label. The workers employ the method of \ac{ls} to train their linear model: worker $n$ determines its empirical loss locally on its dataset using the local \ac{rss}, i.e., 
\begin{align}
F_n\brc{\btheta} = \frac{1}{D_n} \norm{\mX_n \btheta - \by_n}^2, 
\end{align}
for linear model $\btheta \in \setR^{J}$. The global loss is the determined by arithmetic averaging, i.e., 
\begin{align}
F\brc{\btheta} = \frac{1}{N} \sum_{n=1}^N 	F_n\brc{\btheta}.
\end{align}

By straightforward lines of derivation, the global optimum, i.e., the minimizer of $F\brc{\btheta}$, is given by 
\begin{align}
\btheta^\star = \dbc{\sum_{n=1}^N \mX_n^\trp \mX_n}^{-1} \sum_{n=1}^N \mX_n^\trp \by_n.
\end{align}
We use this analytical solution as the reference in our experiment. For this basic setting, we consider deterministic gradient descent, i.e., each worker has a single mini-batch that contains all $500$ data-points. Throughout the simulations, the learning rate is kept fixed at $\eta = 10^{-2}$. 

We generate synthetic data following the approach in \cite{mitra2021linear}. The local datasets are generated independently via a Gaussian linear model: for worker $n$, the data-points are sampled independently from an \ac{iid} zero-mean and unit-variance Gaussian process, i.e., $\bxx_n \sim \man\brc{0, \mI_J}$. To label these data-points, we generate the ground truth model $\bt_n \in \setR^J$ \ac{iid} according to a Gaussian distribution with mean $u_n$ and variance $h^2$, i.e., $\bt_n \sim\man \brc{u_n, h^2 \mI_J}$.  The mean $u_n$ is sampled from a Gaussian process with mean $U$ and variance $\sigma^2$, i.e., $u_n \sim\man \brc{U, \sigma^2}$. The labels are determined via the linear model as 
\begin{align}
	\by_n = \mX_n \bt_n + \be_n,
\end{align}
where $\be_n \sim \man\brc{0, \epsilon^2 \mI_{D_n}}$ form some positive real $\epsilon$. 

\subsubsection*{Numerical Results}
We evaluate the performance by tracking the \textit{optimality gap}: in iteration $t$, we compute the difference between the globally updated model parameter, i.e., $\btheta^t$, and the global minimizer $\btheta^\star$. In other words, in each iteration, we compute the distance
\begin{align}
\delta^t = \norm{\btheta^t - \btheta^\star}.
\end{align}
Recall that the sparsification factor $S$ is defined to be fraction of selected entries at each worker, i.e., $S = k/J$. 

\begin{figure*}
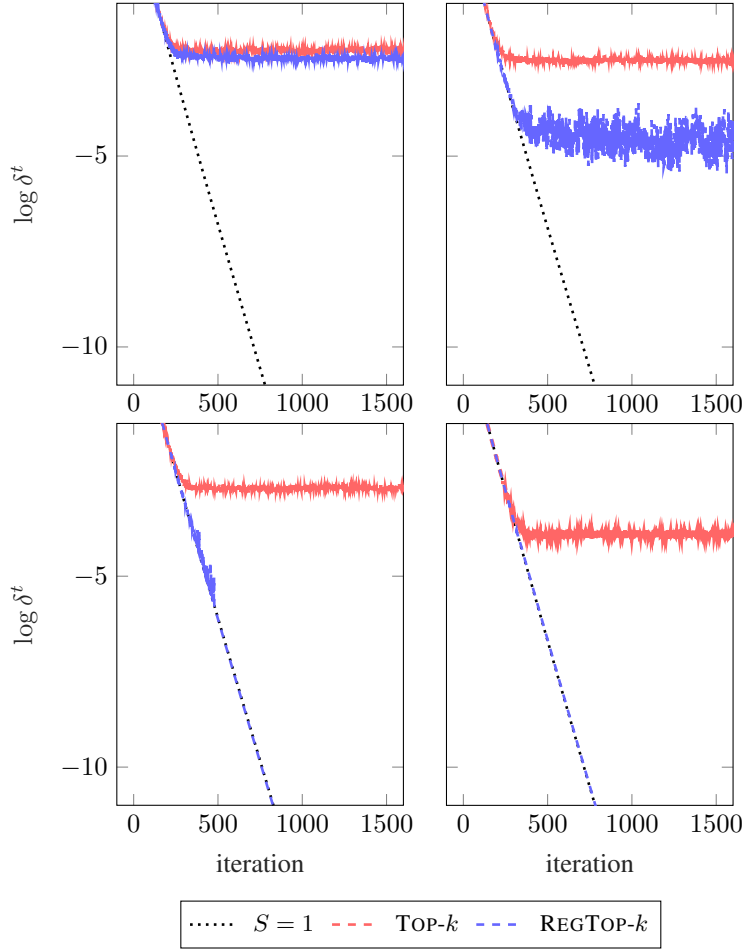

	\begin{center}
		\input{Figures/Reg_top_K_4.tex}
		\input{Figures/Reg_top_K_5.tex}
		
		\input{Figures/Reg_top_K_6.tex}
		\input{Figures/Reg_top_K_9.tex}
	\end{center}\vspace{7mm}
	\caption{Optimality gap vs number of iterations for various sparsity factors $S=0.4$ (top left), $S=0.5$ (top right), $S=0.6$ (bottom left), and $S=0.9$ (bottom right). \rgtpK starts to converge to the global optimum as $S$ surpasses a specific threshold, whereas \tpK keeps converging to a point in the vicinity of the global optimum.}
	\label{fig:2}
\end{figure*}

Figure~\ref{fig:2} sketches the optimality gap $\delta^t$  in logarithmic scale against the number of iterations for \rgtpK, as well as the benchmarks, i.e., \tpK and no sparsification case. Here, we generate the local datasets with $U=0$, $\sigma^2 = 5$, $h^2 = 1$ and $\epsilon^2 = 0.5$. The figure shows the convergence for four sparsity factors, namely $S = 0.4$, $S = 0.5$, $S = 0.6$ and $S=0.9$. As observed, the \rgtpK algorithm starts to track the (non-sparsified) distributed \ac{sgd} at $S=0.6$ while \tpK remains at a certain distance from the optimal solution.  
This behavior can be explained as follows: for both sparsification approaches, the aggregation of large local gradient entries gradually moves the initial point towards the global optimum. At a certain vicinity of the optimal point, the impact of smaller gradient entries in convergence become more dominant. \tpK selects these entries only after large error accumulation, which due to the learning rate scaling leads to oscillation around the optimal point at a fixed distance. To avoid such oscillation, \tpK would need significant damping of the learning rate. \rgtpK, however, selects the small (but dominant) entries at lower error accumulation levels, which prevents large scaling.

\begin{figure}
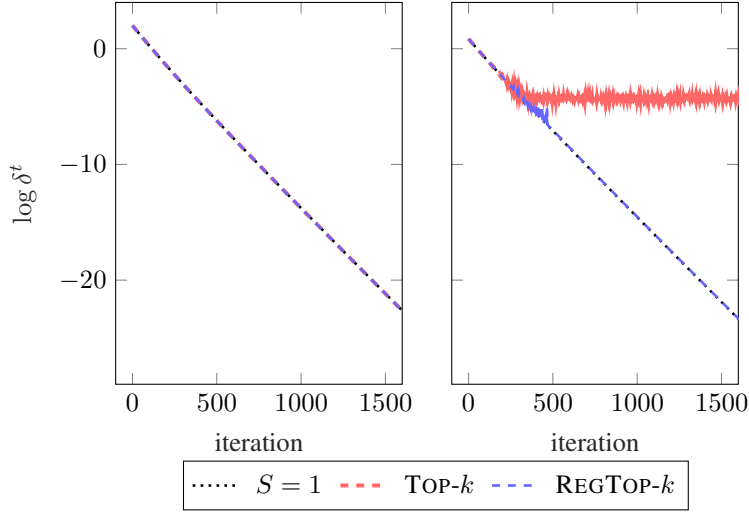

	\begin{center}
		\input{Figures/Reg_top_K_homog.tex}
		\input{Figures/Reg_top_K_hetro1.tex}
	\end{center}\vspace{7mm}
	\caption{Homogeneity (left) vs heterogeneity (right): with heterogeneity, \tpK remains away from the global optimum, whereas \rgtpK converges to~it. }
	\label{fig:hom_het}
\end{figure}

Intuitively, \rgtpK should be more robust against heterogeneity in the network. To investigate this point, we focus on a strictly homogeneous dataset distribution in which the true models of all workers are identical, i.e., $\bt_n = \bt_0$ for all $n \in \dbc{N}$ and the error variance is zero, i.e., $\epsilon = 0$. In this case, the minimizer of the global loss function is identical to that of local ones. Figure~\ref{fig:hom_het} (left) shows $\log \delta^t$ against iterations for \rgtpK and the benchmarks. For \tpK and \rgtpK, the sparsification level is set to $S=0.6$. As the figure shows, both sparsification schemes track the distributed \ac{sgd} curve in the strictly homogeneous setting. This observation is expected, as all the local losses are minimized at the same point. We now introduce heterogeneity to the setting by generating $\bt_n$ independently for each worker with $U=0$, $\sigma^2 = 2$ and $h^2 = 1$. The error variance is further set to $\epsilon^2 = 0.5$. The results are shown in Figure~\ref{fig:hom_het} (right). The \tpK scheme oscillates in this case among points that are in a fixed vicinity of the global optima. This behavior is due to the large learning rate scaling property discussed earlier. 
Unlike \tpK, \rgtpK can converge to the global optimum in both homogeneous and heterogeneous cases.

\begin{figure}
	\begin{center}
		\input{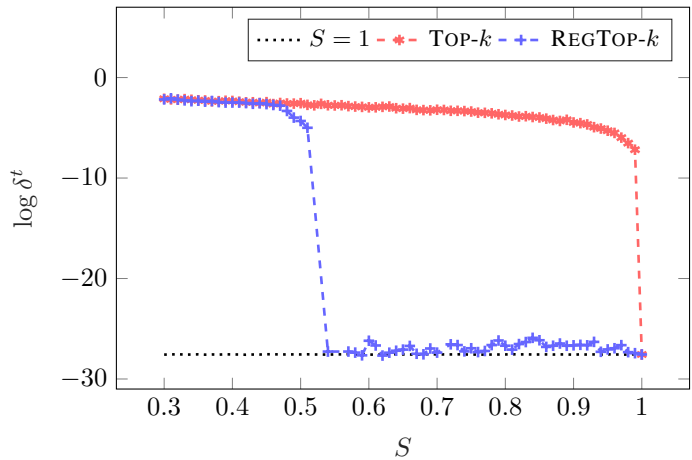}
	\end{center}
	\caption{Optimality gap vs sparsity. \tpK converges to the global optimum only at $S=1$, whereas \rgtpK starts converging at $S=0.55$.}
	\label{fig:del_vs_spsty}
\end{figure}

We next compare the robustness of \rgtpK against \tpK by plotting the optimality gap against at iteration $t=2500$ against the sparsity factor. The results are shown in Figure~\ref{fig:del_vs_spsty}, where the reported optimality gap is averaged over $50$ samples. For this figure, we consider the same heterogeneous setting as the one considered in Figure~\ref{fig:2}. As the figure shows, \tpK converges to the global optimum only when the sparsity reaches $1$. Unlike \tpK, \rgtpK starts converging to the global optimum as the sparsity exceeds $S=0.5$. This is a significant drop demonstrating the robustness achieved by the proposed Bayesian regularization.

\subsection{Classification via ResNet-18}
\begin{figure}
	\begin{center}
		\input{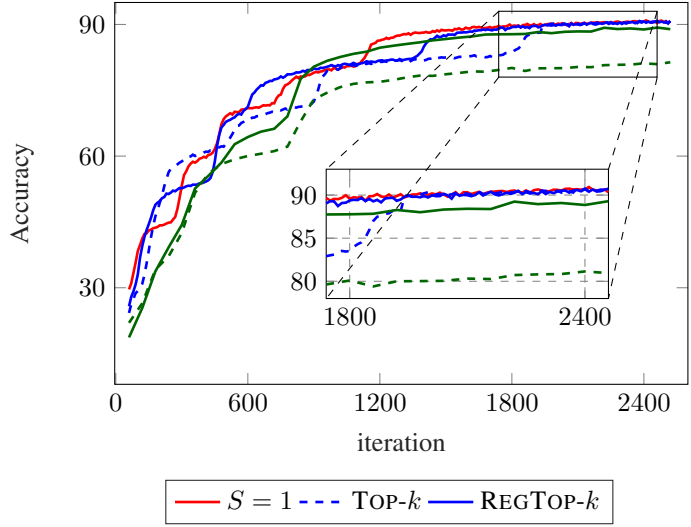}
	\end{center}
	\caption{ResNet-18 on CIFAR-10 with $1\%$ (blue) and $0.1\%$ (green) sparsification. Solid and dashed lines show \rgtpK and \tpK, respectively.}
	\label{fig:resnet}
\end{figure}

We next employ \rgtpK to sparsify the communication carried out in the distributed training of ResNet-18 on CIFAR-10. The training dataset consists of 50,000 images sampled from CIFAR-10 at random. The training samples are uniformly distributed among $N=8$ workers, which compute their local gradients over mini-batches of size $D_n=64$. The aggregation is performed by arithmetic averaging, i.e., $\omega_n= 1/8$. The learning rate is set to $\eta=0.01$ and scheduled during the training. We consider two cases. Namely, the workers sparsify their local gradients with ($i$) $S=0.01$, i.e., $1\%$, and ($ii$) $S=0.001$, i.e., $0.1\%$. Figure~\ref{fig:resnet} shows the test accuracy against the number of rounds for both \tpK and \rgtpK. To keep the comparison fair, we consider the same initialization of the global model for both algorithms, and the batch samplers of the workers are set identical. As the baseline, we further plot the learning curve for training without sparsification, i.e., $S=1$, in red. In Figure~\ref{fig:resnet}, the dashed and solid lines show the results for \tpK and \rgtpK, respectively. As the figure shows, in case ($i$), both \tpK and \rgtpK converge to the baseline. This follows from the fact that $1\%$ sparsification for a large model like ResNet-10 does not impact the performance noticeably. As we reduce the sparsity factor to $0.01\%$ in case ($ii$), we see that after the first 600 iterations, the model trained by \rgtpK sparsification starts to give strictly higher accuracy as compared to the one trained by \tpK. 
This observation validates our initial conclusion that \rgtpK sparsifies more efficiently at higher compression ratios. It further indicates that our earlier findings, illustrated in-depth for the simpler problem of linear regression, extend to a larger scope of problems. 

\subsection{Fine-tuning on ImageNette}
In this section, we evaluate the performance of \rgtpK and compare it against \tpK sparsification on a broader range of computational models. To this end, we fine-tune multiple pretrained models on the \textit{ImageNette} dataset, which is a publicly-available subset of ImageNet with resized samples and is widely used for evaluation of computational models \cite{imagenette,mo2022adversarial}. Throughout the experiments, we use \texttt{ImageNette-160} in which the original images are resized to $160 \times 160$ pixels for computational efficiency. For fine-tuning, we consider the following models pre-trained on the ImageNet dataset \cite{5206848}:
\begin{enumerate}
	\item SqueezeNet, proposed in \cite{iandola2016squeezenet}, is a compact \ac{cnn} architecture designed to achieve AlexNet-level accuracy with significantly fewer parameters, inspired by the architecture of AlexNet \cite{NIPS2012_c399862d}. In our experiments, we use SqueezeNet 1.0, which contains approximately 1.24 million parameters.
	\item ShuffleNetV2, proposed in \cite{ma2018shufflenet} as an improvement to ShuffleNet \cite{zhang2018shufflenet}, which is a lightweight \ac{cnn} architecture designed for high efficiency on mobile devices. In our experiments, we use ShuffleNetV2 (1.0×), which contains approximately 2.3 million parameters.
	\item MobileNetV2, which is proposed in \cite{sandler2018mobilenetv2} introducing inverted residual blocks and linear bottlenecks to the original MobileNet architecture \cite{howard2017mobilenets}. In our experiments, we use the classical MobileNetV2 with approximately 3.4 million parameters. 
	\item EfficientNet, which is proposed in \cite{tan2019efficientnet}. In our experiments, we use EfficientNet-B0, which contains approximately 5.3 million parameters.
	\item ResNet-152 in \cite{he2016deep}, which consists of approximately 60.2 million parameters.
\end{enumerate}

\begin{figure}
	\begin{center}
		\begin{tikzpicture}
  \begin{axis}[
	width=2.7in, 
height=1.5in, 
at={(1.262in,0.7in)},
scale only axis,  
    xlabel={$\mu$},
    ylabel={Accuracy},
    xmin=-.5, xmax=10.5,
    ymin=89.5, ymax=93.5,
    legend pos=north east,
    legend cell align=left,
    cycle list={{orange!80!black,mark=o},{violet!80!black,mark=s}},
  ]
    \addplot[name path=uwgan_upper,draw=none] coordinates {
    (0,90.98)
     (1,92.46)
     (2, 92.73)
     (4,93.01)
     (6,92.94)
     (8,92.94)
     (10,92.94)
    };
    \addplot[name path=uwgan_lower,draw=none] coordinates {
    	(0,90.27)
    	(1,91.82)
    	(2, 91.85)
    	(4, 92.15)
    	(6,91.69)
    	(8,91.69)
    	(10,91.87)
    };
    \addplot[fill=orange!20, draw=none] fill between[
      of=uwgan_upper and uwgan_lower
    ];
    \addplot[very thick,forget plot] coordinates {
    (0,90.81)
    (1,92.17)
    (2, 92.38)
    (4,92.62)
    (6,92.47)
    (8,92.46)
    (10,92.45)
    };
   
  \end{axis}
\end{tikzpicture}
	\end{center}
	\caption{Tuning hyperparameter $\lambda$ in \rgtpK for MobileNetV2 at sparsity $0.1\%$. The point $\mu=0$ corresponds to \tpK.}
	\label{fig:ablation}
\end{figure}
For these experiments, we consider a dense network with $N=20$ workers. The training samples are distributed uniformly among the workers. Each worker deploys a distributed version of the Adam optimizer \cite{kingma2014adam} and applies \rgtpK for gradient sparsification. We consider two sparsity levels: ($i$) $1\%$ sparsity ($S=0.01$) and ($ii$) $0.1\%$ sparsity ($S=0.001$). The accuracy and loss for both sparsity levels are evaluated on the validation set after $8$ epochs of fine-tuning and compared against the \tpK sparsification method. All experiments are conducted over $10$ common random seeds for each model and sparsity level, with the mean and standard deviation of the results reported in Table~\ref{table:FineTune}. Throughout the experiments, we tune the hyperparameter $\mu$ for each model by grid-search over the interval $\mu \in [1,10]$. Fig.~\ref{fig:ablation} shows a sample curve for MobileNetV2 with $0.1\%$ sparsification, where the accuracy of the fine-tuned model is plotted against $\mu$. Note that in this plot $\mu= 0$ shows the result for the \tpK algorithm.\footnote{Recall from Section~\ref{sec:RegTopK} that \rgtpK reduces to \tpK as $\mu \to 0$.} From the figure, one can observe that \rgtpK is rather stable against changes in $\mu$. The numerical results suggest that tuning of $\mu$ can help reduce performance variance.

\begin{table*}[t]
	\caption{
		\rgtpK vs \tpK: 
		{\normalfont Fine-tuning various models on ImageNette \cite{imagenette} for two sparsity levels. The results for \rgtpK show statistically significant improvement over \tpK with $p$-values less than $0.01$. At the lower sparsity level  ($0.01\%$ sparse), the accuracy and loss degradation observed with \rgtpK is significantly less pronounced compared with \tpK.
		}
	}
	\centering\scriptsize
	\begin{tabular}[t]{|c|cc|c|}
		\hline
		& & $1\%$ sparse &$0.1\%$ sparse   \\
		\hline
		Model & 
		&\begin{tabular}[t]{ll}
			Accuracy$\qquad\qquad$  &Loss$\;\;\;$
		\end{tabular}
		&\begin{tabular}[t]{ll}
			Accuracy$\qquad\qquad$  &Loss$\;\;\;$
		\end{tabular} \\
		\hline
		SqueezeNet 
		& \tpK 
		&\begin{tabular}[t]{ll}
			$82.73\pm 0.40\%$  &$0.5639\pm 0.0087$
		\end{tabular}
		&\begin{tabular}[t]{ll}
			$74.57 \pm 1.10\%$  &$0.8661 \pm 0.0282$
		\end{tabular} \\
		& \rgtpK 
		&\begin{tabular}[t]{ll}
			$87.21\pm 0.45 \%$  &$0.3867 \pm 0.0073$
		\end{tabular} 
		&\begin{tabular}[t]{ll}
			$82.14 \pm 0.66\%$  &$0.5415 \pm 0.0212$
		\end{tabular}\\
		\hline
		ShuffleNetV2
		& \tpK 
		&\begin{tabular}[t]{ll}
			$74.03\pm 1.09\%$  &$2.0905\pm 0.0021$
		\end{tabular}
		&\begin{tabular}[t]{ll}
			$51.63 \pm 4.32\%$  &$2.2567 \pm 0.0019$
		\end{tabular} \\
		& \rgtpK 
		&\begin{tabular}[t]{ll}
			$82.81\pm 0.19 \%$  &$1.5050 \pm 0.0016$
		\end{tabular} 
		&\begin{tabular}[t]{ll}
			$81.83 \pm 0.36\%$  &$1.6085 \pm 0.0015$
		\end{tabular}\\
		\hline
		MobileNetV2
		& \tpK 
		&\begin{tabular}[t]{ll}
			$92.28\pm 0.19\%$  &$0.2279\pm 0.0035$
		\end{tabular}
		&\begin{tabular}[t]{ll}
			$90.81 \pm 0.34\%$  &$0.3521 \pm 0.0045$
		\end{tabular} \\
		& \rgtpK 
		&\begin{tabular}[t]{ll}
			$93.09\pm 0.15 \%$  &$0.1836 \pm 0.0021$
		\end{tabular} 
		&\begin{tabular}[t]{ll}
			$92.60 \pm 0.33\%$  &$0.2006 \pm 0.0131$
		\end{tabular}\\
		\hline
		EfficientNet 
		& \tpK 
		&\begin{tabular}[t]{ll}
			$88.42\pm 0.19\%$  &$0.4153\pm 0.0055$
		\end{tabular}
		&\begin{tabular}[t]{ll}
			$85.56 \pm 0.39\%$  &$0.7539 \pm 0.0064$
		\end{tabular} \\
		& \rgtpK 
		&\begin{tabular}[t]{ll}
			$90.05\pm 0.06 \%$  &$0.2805 \pm 0.0049$
		\end{tabular} 
		&\begin{tabular}[t]{ll}
			$89.93 \pm 0.15\%$  &$0.2914 \pm 0.0067$
		\end{tabular}\\
		\hline
		ResNet-152
		& \tpK 
		&\begin{tabular}[t]{ll}
			$93.99\pm 0.18\%$  &$0.1480\pm 0.0034$
		\end{tabular}
		&\begin{tabular}[t]{ll}
			$92.81 \pm 0.14\%$  &$0.2354 \pm 0.0041$
		\end{tabular} \\
		& \rgtpK 
		&\begin{tabular}[t]{ll}
			$94.33 \pm 0.19 \%$  &$0.1307 \pm 0.0114$
		\end{tabular} 
		&\begin{tabular}[t]{ll}
			$94.76 \pm 0.22\%$  &$0.1193 \pm 0.0100$
		\end{tabular}\\
		\hline
	\end{tabular}
	\label{table:FineTune}
\end{table*}

The results in Table~\ref{table:FineTune} demonstrate consistent improvements in validation metrics for all models across both sparsity levels. To confirm that these improvements are \textit{statistically significant}, we conducted \textit{paired $t$-tests} and \textit{Wilcoxon} signed-rank tests on the validation results \cite{hollander2013nonparametric}. All experiments yielded $p$-values below the $0.01$ threshold, indicating statistically significant improvements of \rgtpK over \tpK. Furthermore, the results show that at the lower sparsity level (corresponding to a higher compression rate) \tpK experiences substantial performance degradation, whereas \rgtpK exhibits only marginal decline. These findings align with our earlier observations on linear regression and ResNet-18 classification tasks, validating the conjecture that insights from the simpler problems extend to a broad range of distributed learning settings.

\section{Conclusions}
Using information collected during training, local gradients can be more efficiently sparsified. We have invoked this idea to develop a Bayesian framework for sparsification. The proposed scheme regularizes \tpK sparsification to control its learning rate scaling. Numerical investigations validate our derivations: \rgtpK can track the performance of non-sparsified distributed learning at significantly lower sparsity factors than \tpK. Interestingly, this gain is achieved with no considerable increase in computation complexity. The scope of this work can be extended in various respects. Most naturally, the proposed Bayesian regularization can be developed for other notions of optimality studied in \cite{sahu2021rethinking}. 


\bibliographystyle{IEEEtran}
\bibliography{ref}

\section*{Appendices}
\appendix

\section{Approximating $I_i \brc{\bzz_n^{t-1}}$}
\label{append:B}

As in Proposition~\ref{prop:1}, we assume that the entries of the innovation term $ \bxi_n^{t}$ are zero-mean, independent, and distributed according to the same density with different variances. To capture the scaling at the workers, we set the variance to be proportional to the amplitude of the local accumulated gradient. We then write
\begin{align}
	I_{i} \brc{\bzz_n^{t-1} }  
	&= \int_{\setF_i^k } \prod_{j=1}^{J}  p_j \brc{ {x_j- \omega_n a\id{j}{n}^t - z\id{j}{n}^{t-1} } } \dif x_j  ,
\end{align}
where $p_j\brc{\cdot}$ denotes the distribution of $\xi\id{j}{n}^t$. As in Result~\ref{prop:2}, we further assume $p_j\brc{\cdot}$ is symmetric around $0$, i.e., 
\begin{align}
	\int_x^\infty p_j \brc{u} \dif u = \int_{-\infty}^{-x} p_j \brc{u} \dif u
\end{align}
for $x\in \setR$. To keep the notation simple, let us further define 
\begin{align}
	\bar{a}\id{j}{n}^t = 
	 \omega_n a\id{j}{n}^t + z\id{j}{n}^{t-1} .
\end{align}
Note that entry $\bar{a}\id{j}{n}^t$ is known at worker $n$ and is an estimator of the unknown global accumulated gradient entry ${a}\id{j}{}^t$. Using this notation, we can compactly write
\begin{align}
	I_{i} \brc{\bzz_n^{t-1} }  
	&= \int_{\setF_i^k } \prod_{j=1}^{J}  p_j \brc{ x_j - 	\bar{a}\id{j}{n}^t } \dif x_j  .\label{eq:Iz}
\end{align}
The exact calculation of this term is intractable. We can however invoke the method of types to approximate it in the limit of $J\to\infty$ \cite{csiszar1998method}. 

Let us first define the set $\setL_i$ to be the index set of entries in $\bar{\baa}_n^t$ that are smaller than the particular entry $\bar{a}\id{i}{n}^t$, i.e., 
\begin{align}
	\setL_i &= \set{j:    \abs{\bar{a}\id{i}{n}^t} > \abs{\bar{a}\id{j}{n}^t}}. 
\end{align}
We now define the set $\tilde{\setF}_i$ to be the set of points in $\setR^J$ that have identical sorting, i.e., 
\begin{align}
	 \tilde{\setF}_i = \{ \bx\in\setR^J: & \abs{x_i} > \abs{x_j} \; \text{ for } j \in \setL_i  \text{and } \; \abs{x_i} \leq \abs{x_j} \; \text{ for } \; j \notin \setL_i 	
	\}.
\end{align}
The following lemma gives two key properties of $\tilde{\setF}_i$ and its connection to $\setF_i$.

\begin{lemma}
	\label{lem:1}
	The sets $\setL_i$ and $\tilde{\setF}_i$ satisfy the following properties:
\begin{itemize}
	\item[(a)] If $\abs{\setL_i} < J-k$, then $\tilde{\setF}_i$ and $\setF_i^k$ are distinct.
	\item[(b)] If $\abs{\setL_i} \geq J-k$, then $\tilde{\setF}_i \subseteq \setF_i^k$.
\end{itemize}
\end{lemma}

\begin{proof}
	The proof readily follows the definition of $\tilde{\setF}_i$: when $\abs{\setL_i} < J-k$, $\tilde{\setF}_i$ includes only those points in $\setR^J$ for which $x_i$ is not among the top $k$ entries, whereas ${\setF}_i$ includes points whose $i$-th entry is among the top $k$. Hence, $\tilde{\setF}_i \cap \setF_i^k = \emptyset$. This concludes (a). On the other hand, with $\abs{\setL_i} \geq J-k$, $\tilde{\setF}_i$ includes points whose $i$-th entry is among the top $k$ entries. Nevertheless, these points are not the only points with this property, as there are other valid sorting orders for which $x_i$ is among top $k$. We can hence conclude that in this case $\tilde{\setF}_i \subseteq \setF_i^k$ which concludes (b).
\end{proof}

The method of types indicates that with an innovation whose density decays very fast with its amplitude, e.g., a Gaussian innovation, the integrand in the definition of $I_i \brc{\bzz_n^{t-1} }$ is dominated with the type set whose sorting order is identical to that of $\bar{\baa}_n^t$. Consequently, the term $I_i\brc{\bzz_n^{t-1} }$ is close to zero, if $\bar{a}\id{i}{n}^t$ is not among the top $k$ entries of $\bar{\baa}_n^t$. This statement can be illustrated more comprehensively via a large deviations argument \cite{dembo2009large}: as $J$ grows large, the integrand in \eqref{eq:Iz} is dominated over the set $\tilde{\setF}_i$.  From Lemma~\ref{lem:1}, we observe that when $\abs{\setL_i} \geq J-k$, this set lies within $\setF_i^k$ and hence $I\brc{\bzz_n^{t-1}}$ asymptotically tends to the integral over $\tilde{\setF}_i$. On the other hand, in the case of $\abs{\setL_i} < J-k$, the dominant set falls out of $\setF_i^k$, and hence $I\brc{\bzz_n^{t-1}}$ tends to zero. 

Recall the assumption in Result~\ref{prop:2}, which indicates 
	that for a given $j\in \dbc{J}$ and small $\delta$, there exists a small $\varepsilon$, such that 
	\begin{align}
		\int_{-\varepsilon \abs{a\id{j}{n}^t}}^{\varepsilon \abs{a\id{j}{n}^t} } p_j \brc{\xi} \dif \xi \geq 1 - \delta. \label{eq:asp1}
	\end{align}
Considering this assumption, we invoke the large deviations argument and write 
\begin{align}
	I_i \brc{\bzz_n^{t-1} }  
\approx
	 \mathds{1} \set{ \abs{\setL_i} \geq J-k}
	\int_{\tilde{\setF}_i } \prod_{j=1}^{J}  p_j \brc{ x_j - 	\bar{a}\id{j}{n}^t } \dif x_j, 
\end{align}
where we use the notation $\approx$ to indicate that it is an asymptotically accurate approximation. We next note that the constraint $\abs{\setL_i} \geq J-k$ is equivalent to $\bar{\baa}_n^t \in \setF_i^k$. Therefore, we write 
\begin{align}
	I_i \brc{\bzz_n^{t-1} }  
	&\approx \mathds{1} \set{ \bar{\baa}_n^t \in \setF_i^k }
	\int_{\tilde{\setF}_i } \prod_{j=1}^{J}  p_j \brc{ x_j - 	\bar{a}\id{j}{n}^t } \dif x_j.
\end{align}

Let us now consider an expansion of $\tilde{\setF}_i$ for which the upper sorting is dropped. Define 
\begin{align}
	\setP_i = \set{ \bx\in\setR^J: \abs{x_i} > \abs{x_j} \; \text{ for } j \in \setL_i 
	}.
\end{align}
Note that $\setP_i$ is a superset of the dominant set $\tilde{\setF}_i$, and hence is consistent with the large deviations argument. We then write 
\begin{align}
	I_i \brc{\bzz_n^{t-1} }  
	&\approx \mathds{1} \set{ \bar{\baa}_n^t \in \setF_i^k }
	\int_{\setP_i } \prod_{j=1}^{J}  p_j \brc{ x_j - 	\bar{a}\id{j}{n}^t } \dif x_j.
\end{align}
\begin{remark}
	It is worth mentioning that the large deviations argument can be directly applied through defining type sets $\set{\setP_i}$. Nevertheless, we presented the argument with the help of sets $\{\tilde{\setF}_i\}$ for the sake of clarity.
\end{remark}

Considering the definition of $\setP_i$, we can expand the integral over the dominant set $\setP_i$ as 
	\begin{align}
	I_i^0 \brc{\bzz_n^{t-1} }  &= \int_{{\setP}_i } \prod_{j=1}^{J} p_j \brc{ {x_j- \bar{a}\id{j}{n}^t } } \dif x_j  \\
	&= \int	 \prod_{j\in \setL_i}   \int_{-\abs{x_i}}^{\abs{x_i}} p_j \brc{ {x_j- \bar{a}\id{j}{n}^t } } \dif x_j 
	p_i \brc{ {x_i- \bar{a}\id{i}{n}^t } } \dif x_i. 
		\label{eq:I_i0}
\end{align}
Denoting the cumulative distribution of $\xi\id{j}{n}^t$ by $F_j\brc{\cdot}$, we can simplify the integration over $x_j$ for $j\neq i$ as 
\begin{align}
	  \int_{-\abs{x_i}}^{\abs{x_i}} &p_j \brc{ {x_j- \bar{a}\id{j}{n}^t } } \dif x_j = Q_j\brc{\abs{x_i}}, 
\end{align}
where we define $Q_j\brc{\abs{x_i}}$ in terms of $F_j\brc{\cdot}$ as
\begin{align}
	Q_j\brc{\abs{x_i}} = 
	F_j \brc{ \abs{x_i} - \bar{a}\id{j}{n}^t   }  
	- F_j \brc{ -\abs{x_i} - \bar{a}\id{j}{n}^t  }.
\end{align}
From \eqref{eq:asp1}, we can conclude that for a small $\varepsilon$ there exist small $\delta_1$ and $\delta_2$, such that
\begin{align}
	\begin{cases}
		Q_j\brc{\abs{x_i}} \geq 1-\delta_1 &\text{if } \; \abs{x_i} - \abs{\bar{a}\id{j}{n}^t } > \varepsilon \abs{{a}\id{j}{n}^t }\\
		Q_j\brc{\abs{x_i}} < \delta_2 &\text{if } \; \abs{x_i} - \abs{\bar{a}\id{j}{n}^t } < - \varepsilon \abs{{a}\id{j}{n}^t }
	\end{cases}.
\end{align}
Intuitively, this term rapidly changes from zero to one in the vicinity of $\abs{\bar{a}\id{i}{n}^t }$. The slope of this transition is proportional to $ \abs{{a}\id{j}{n}^t }$. Using the definition of $Q_j\brc{\abs{x_i}}$ we have
\begin{align}
I_i^0 \brc{\bzz_n^{t-1} } 
	&= \int	 \prod_{j \in \setL_i}   Q_j\brc{\abs{x_i}}
	p_i \brc{ {x_i- \bar{a}\id{i}{n}^t } } \dif x_i .
\end{align}

To proceed with the derivation, we now invoke a heuristic argument: consider the following product term:
\begin{align}
Q\brc{\abs{x_i}} = 	 \prod_{j\in \setL_i}   Q_j\brc{\abs{x_i}},
\end{align}
which describes the product of $\abs{\setL_i}$ smooth thresholders\footnote{We refer to $Q_j\brc{\abs{x_i}}$ as smooth threshoder, as it approximates hard-thresholding operation at $\abs{\bar{a}\id{i}{n}^t }$ by a smoothed increasing function. Note that it is a \textit{double-sided} thresholder, since it depends on $\abs{x_i}$.} with the threshold of all functions being below $\abs{\bar{a}\id{i}{n}^t}$. We further note that none of these functions depend on $\abs{a\id{i}{n}^t}$. We hence approximate $Q\brc{{x_i}}$ with a hard thresholding function at a point $T \leq \abs{\bar{a}\id{i}{n}^t}$ that does not scale with $\abs{a\id{i}{n}^t}$. With this heuristic approximation we can write
\begin{subequations}
	\begin{align}
I_i^0 \brc{\bzz_n^{t-1} }   &\approx
\int_{-\infty}^{-T}
p_i \brc{ {x_i- \bar{a}\id{i}{n}^t } } \dif x_i  + \int_T^\infty
p_j \brc{ {x_i- \bar{a}\id{i}{n}^t } } \dif x_i \\
&=
 \Pr\set{ \abs{\bar{a}\id{i}{n}^t + \xi\id{i}{n}^t} > T}.
\end{align}
\end{subequations}
Consequently, we write 
\begin{align}
	I_i \brc{\bzz_n^{t-1} }  
	&\approx \mathds{1} \set{ \bar{\baa}_n^t \in \setF_i^k }
	 \Pr\set{ \abs{\bar{a}\id{i}{n}^t + \xi\id{i}{n}^t} > T},
\end{align}
for some $T$ that is fixed in $\abs{a\id{i}{n}^t}$, and a random variable $\xi\id{i}{n}^t$ whose variance scales with $\abs{a\id{i}{n}^t}$. 

\begin{remark}
	Following the heuristics mentioned above, one can directly start from the last expression to heuristically calculate the likelihood. We however have given the above discussions to illustrate how we have arrived at this conclusion. It is worth mentioning that even by specifying the distribution of innovation, derivation of the likelihood is not trivial. This necessitates the use of heuristics.
\end{remark}

We now try to understand how $I_i \brc{\bzz_n^{t-1} }$ scales. To this~end, recall the definition of  ${\Delta\id{i}{n}^t}$ in \eqref{eq:Delta_def}, which can be alternatively written as
\begin{align}
	{\Delta\id{i}{n}^t} = \frac{z\id{i}{n}^{t-1}}{\omega_n a\id{i}{n}^t}.
\end{align}
Focusing on top $k$ entries of $\bar{\baa}_n^t$, we can write that 
\begin{subequations}
	\begin{align}
	I_i \brc{\bzz_n^{t-1} }   &\approx
	\Pr\set{ \abs{\bar{a}\id{i}{n}^t + \xi\id{i}{n}^t} > T} \\
	&=  \Pr\set{\omega_n\abs{a\id{i}{n}^t} \left\vert 1+\Delta\id{i}{n}^t + \bar{\xi}^t_i \right\vert > T },\label{eq:ineq1}
\end{align}
\end{subequations}
where we define $\bar{\xi}_i$ to be
\begin{align}
	\bar{\xi}_i^t = \frac{\xi\id{i}{n}^t}{\omega_n a\id{i}{n}^t} .
\end{align}

\subsection{Scaling with $\abs{a\id{i}{n}^t}$}
We first note that due to the scaling of $\xi\id{i}{n}^t$, the variance of $\bar{\xi}_i^t$ does not depend on $\abs{a\id{i}{n}^t}$. This indicates that $I_i \brc{\bzz_n^{t-1} }$ increases monotonically with $\abs{a\id{i}{n}^t}$. Furthermore, 
noting that both terms in the two sides of the inequality inside the argument of \eqref{eq:ineq1} are non-negative, 
we can use Markov's inequality and write
\begin{subequations}
	\begin{align}
		I_i \brc{\bzz_n^{t-1} }   &\approx
 \Pr\set{\omega_n\abs{a\id{j}{n}^t} \left\vert 1+\Delta\id{j}{n}^t + \bar{\xi}_j^t \right\vert > T }\\
&\leq 
\frac{1}{T} \Ex{ 
\omega_n\abs{a\id{j}{n}^t} \left\vert 1+\Delta\id{j}{n}^t + \bar{\xi}_j^t \right\vert 
 }{}\\
&= 
\frac{\abs{a\id{j}{n}^t}}{T} \Ex{ \omega_n  \left\vert 1+\Delta\id{j}{n}^t + \bar{\xi}_j^t \right\vert  }{}.\label{eq:final1}
\end{align}
\end{subequations}
The upper bound indicates that the growth of $I_i \brc{\bzz_n^{t-1} }$ with $\abs{a\id{i}{n}^t}$ is at most linear. This is consistent with the \tpK prior, given in Definition~\ref{def:Prior}. 

\subsection{Scaling with $\Delta\id{i}{n}^t$}
We next investigate the scaling with $\Delta\id{i}{n}^t$. To this end, we write
\begin{subequations}
	\begin{align}
	I_i \brc{\bzz_n^{t-1} }   \approx&
  \Pr\set{\omega_n\abs{a\id{i}{n}^t} \left\vert 1+\Delta\id{i}{n}^t + \bar{\xi}^t_i \right\vert > T }\\
	 =& 
	\Pr\set{  \bar{\xi}_i^t > \tau_i - {1-\Delta\id{i}{n}^t} } +
	\Pr\set{ \bar{\xi}_i^t < -\tau_i - {1-\Delta\id{i}{n}^t} },
\end{align}
\end{subequations}
where we define $\tau_i > 0$ as
\begin{align}
	\tau_i =  \frac{T}{\omega_n\abs{a\id{i}{n}^t} }.
\end{align}
Following the symmetry assumption on $p_i\brc{\cdot}$, we can alternatively write
\begin{align}
	I_i \brc{\bzz_n^{t-1} }   \approx&	
	\Pr\set{  \bar{\xi}_i^t <  \abs{1+\Delta\id{i}{n}^t} - \tau_i } +
	\Pr\set{ \bar{\xi}_i^t > \tau_i  + \abs{1+\Delta\id{i}{n}^t} }.
\end{align}
The fast decaying assumption on $p_i\brc{\cdot}$ indicates that the right hand side of the above equation is dominated by the first term. 
We now assume that the random variable $\bar{\xi}_i^t + \tau_i$ is approximately identically distributed over $i$. Let $u_\mu \brc{\cdot}$ denote its cumulative distribution with $\mu > 0$ being a parameter specifying the distribution, e.g., the variance. 
Note that $u_\mu \brc{x}$ is an increasing function in $\abs{x}$ whose minimum occurs at $x=0$. Using this notation, we can finally write
\begin{align}
	I_i\brc{\bzz_n^{t-1}} \propto \abs{a\id{i}{n}^t} u_{\mu} \brc{\abs{1+\Delta\id{i}{n}^t}}.
\end{align}

\begin{remark}
	\label{rmk:3}
	Considering \eqref{eq:final1}, one may assume a sub-linear scaling, e.g., scaling of order $\abs{a\id{i}{n}^t}^y$ for some $y\leq 1$. This	scaling suggests that in \rgtpK sparsification, the selection metric for entry $i$ should be set to $m_i$, where 
	\begin{align}
		m_i = \abs{a\id{i}{n}^t}^y u_\mu \brc{{\abs{1+ \Delta\id{i}{n}^t}} }
	\end{align}
	with $y \leq 1$. Here $y$, similar to $\mu$, is a tunable hyper-parameter. By dropping the regularization, i.e., reducing the sparsification to standard \tpK, this hype-parameter can also be neglected, since it does not impact the sorting anymore. 
\end{remark}

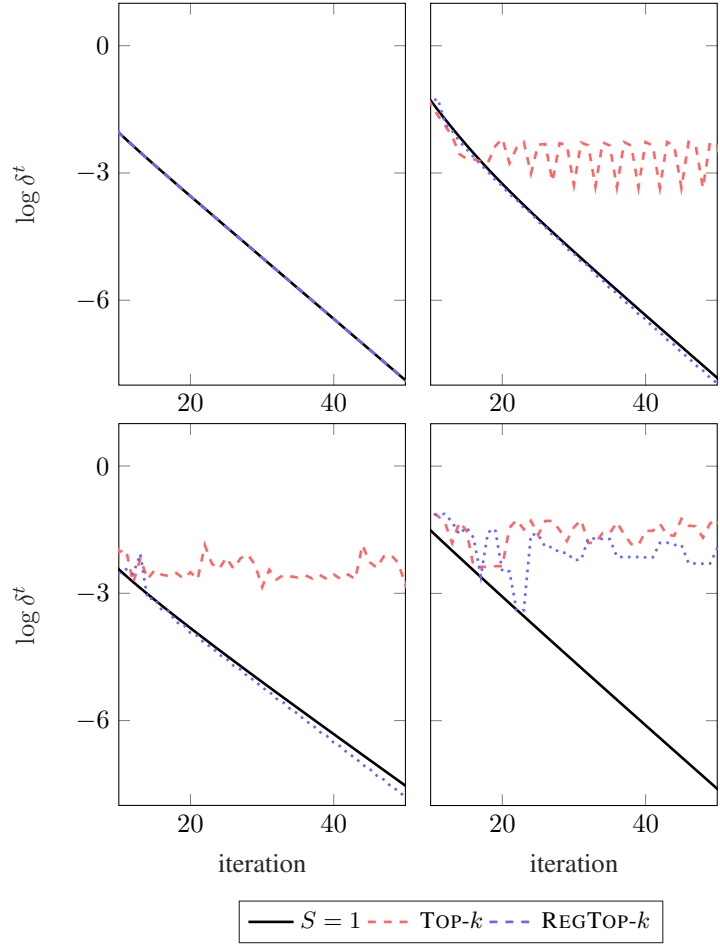
\begin{figure*}
	\begin{center}
%
%
\begin{tikzpicture}
	
	\begin{axis}[%
	width=1.5in, 
height=2in, 
		at={(1.262in,0.7in)},
		scale only axis,  
		xmin=10,  
		xmax=50,  
		xtick={0, 20,  40,  60}  ,
		xticklabels={{$0$},{$20$}, {$40$},  {$60$}},
		xlabel style={font=\color{white!15!black}},  
		ymin=-8,  
		ymax=1,  
		ytick={0  , -3, -6}  ,
		yticklabels={{$0$} , {$-3$}, {$-6$}}  ,
		ylabel style={font=\color{white!15!black}}, 
		ylabel={$\log \delta^t$},
		yminorticks=true,
		axis background/.style={fill=white},
		legend columns = 3,
		legend style={at={(1.9, -1.35)}, legend cell align=left, align=left, draw=white!15!black,fill=none,overlay},
		]
		\addplot[color=black, line width=1.0pt]
		table[row sep=crcr]{%
		1    -0.33416800045006007 \\
		2    -0.5591904975570117 \\
		3    -0.774980573618763 \\
		4    -0.981270203977436 \\
		5    -1.1781924704259892 \\
		6    -1.366260171208437 \\
		7    -1.5462858900217962 \\
		8    -1.7192685108151877 \\
		9    -1.8862765881218588 \\
		10    -2.048352060888621 \\
		11    -2.2064452152670557 \\
		12    -2.361380614716404 \\
		13    -2.5138473373981483 \\
		14    -2.664405100199566 \\
		15    -2.8134988944545802 \\
		16    -2.9614768416732002 \\
		17    -3.108608016506269 \\
		18    -3.255098541009125 \\
		19    -3.401105275273749 \\
		20    -3.5467470202812637 \\
		21    -3.6921134460686065 \\
		22    -3.8372720766876056 \\
		23    -3.9822736833843955 \\
		24    -4.127156409463349 \\
		25    -4.2719489035460425 \\
		26    -4.416672687807405 \\
		27    -4.5613439415609935 \\
		28    -4.70597484106984 \\
		29    -4.850574564166364 \\
		30    -4.9951500425941076 \\
		31    -5.139706524960827 \\
		32    -5.284247997774835 \\
		33    -5.428777500273994 \\
		34    -5.573297359841727 \\
		35    -5.717809368078654 \\
		36    -5.862314912541389 \\
		37    -6.006815075368039 \\
		38    -6.151310707170181 \\
		39    -6.295802482447655 \\
		40    -6.440290941196843 \\
		41    -6.584776520198065 \\
		42    -6.729259576584833 \\
		43    -6.873740405635768 \\
		44    -7.01821925424189 \\
		45    -7.162696331129913 \\
		46    -7.30717181465324 \\
		47    -7.4516458587520775 \\
		48    -7.5961185975410235 \\
		49    -7.7405901488544915 \\
		50    -7.885060617009411 \\
		51    -8.029530094975112 \\
		52    -8.173998666088169 \\
		53    -8.318466405421713 \\
		54    -8.462933380893864 \\
		55    -8.607399654168583 \\
		56    -8.751865281398437 \\
		57    -8.896330313845532 \\
		58    -9.040794798405125 \\
		59    -9.185258778049938 \\
		60    -9.329722292208874 \\
		};
		\addlegendentry{\small $S=1$}
		
		\addplot[color=red!60, dashed, line width=1.0pt]
		table[row sep=crcr]{%
		1    -0.33416800045006007 \\
		2    -0.5591904975570117 \\
		3    -0.774980573618763 \\
		4    -0.981270203977436 \\
		5    -1.1781924704259892 \\
		6    -1.366260171208437 \\
		7    -1.5462858900217962 \\
		8    -1.7192685108151877 \\
		9    -1.8862765881218588 \\
		10    -2.048352060888621 \\
		11    -2.2064452152670557 \\
		12    -2.361380614716404 \\
		13    -2.5138473373981483 \\
		14    -2.664405100199566 \\
		15    -2.8134988944545802 \\
		16    -2.9614768416732002 \\
		17    -3.108608016506269 \\
		18    -3.255098541009125 \\
		19    -3.401105275273749 \\
		20    -3.5467470202812637 \\
		21    -3.6921134460686065 \\
		22    -3.8372720766876056 \\
		23    -3.9822736833843955 \\
		24    -4.127156409463349 \\
		25    -4.2719489035460425 \\
		26    -4.416672687807405 \\
		27    -4.5613439415609935 \\
		28    -4.70597484106984 \\
		29    -4.850574564166364 \\
		30    -4.9951500425941076 \\
		31    -5.139706524960827 \\
		32    -5.284247997774835 \\
		33    -5.428777500273994 \\
		34    -5.573297359841727 \\
		35    -5.717809368078654 \\
		36    -5.862314912541389 \\
		37    -6.006815075368039 \\
		38    -6.151310707170181 \\
		39    -6.295802482447655 \\
		40    -6.440290941196843 \\
		41    -6.584776520198065 \\
		42    -6.729259576584833 \\
		43    -6.873740405635768 \\
		44    -7.01821925424189 \\
		45    -7.162696331129913 \\
		46    -7.30717181465324 \\
		47    -7.4516458587520775 \\
		48    -7.5961185975410235 \\
		49    -7.7405901488544915 \\
		50    -7.885060617009411 \\
		51    -8.029530094975112 \\
		52    -8.173998666088169 \\
		53    -8.318466405421713 \\
		54    -8.462933380893864 \\
		55    -8.607399654168583 \\
		56    -8.751865281398437 \\
		57    -8.896330313845532 \\
		58    -9.040794798405125 \\
		59    -9.185258778049938 \\
		60    -9.329722292208874 \\
		};
		\addlegendentry{\small \tpK}
		
		\addplot[color=blue!60, dashed, line width=1.0pt]
		table[row sep=crcr]{%
		1    -0.33416800045006007 \\
		2    -0.5591904975570117 \\
		3    -0.774980573618763 \\
		4    -0.981270203977436 \\
		5    -1.1781924704259892 \\
		6    -1.366260171208437 \\
		7    -1.5462858900217962 \\
		8    -1.7192685108151877 \\
		9    -1.8862765881218588 \\
		10    -2.048352060888621 \\
		11    -2.2064452152670557 \\
		12    -2.361380614716404 \\
		13    -2.5138473373981483 \\
		14    -2.664405100199566 \\
		15    -2.8134988944545802 \\
		16    -2.9614768416732002 \\
		17    -3.108608016506269 \\
		18    -3.255098541009125 \\
		19    -3.401105275273749 \\
		20    -3.5467470202812637 \\
		21    -3.6921134460686065 \\
		22    -3.8372720766876056 \\
		23    -3.9822736833843955 \\
		24    -4.127156409463349 \\
		25    -4.2719489035460425 \\
		26    -4.416672687807405 \\
		27    -4.5613439415609935 \\
		28    -4.70597484106984 \\
		29    -4.850574564166364 \\
		30    -4.9951500425941076 \\
		31    -5.139706524960827 \\
		32    -5.284247997774835 \\
		33    -5.428777500273994 \\
		34    -5.573297359841727 \\
		35    -5.717809368078654 \\
		36    -5.862314912541389 \\
		37    -6.006815075368039 \\
		38    -6.151310707170181 \\
		39    -6.295802482447655 \\
		40    -6.440290941196843 \\
		41    -6.584776520198065 \\
		42    -6.729259576584833 \\
		43    -6.873740405635768 \\
		44    -7.01821925424189 \\
		45    -7.162696331129913 \\
		46    -7.30717181465324 \\
		47    -7.4516458587520775 \\
		48    -7.5961185975410235 \\
		49    -7.7405901488544915 \\
		50    -7.885060617009411 \\
		51    -8.029530094975112 \\
		52    -8.173998666088169 \\
		53    -8.318466405421713 \\
		54    -8.462933380893864 \\
		55    -8.607399654168583 \\
		56    -8.751865281398437 \\
		57    -8.896330313845532 \\
		58    -9.040794798405125 \\
		59    -9.185258778049938 \\
		60    -9.329722292208874 \\
		};
		\addlegendentry{\small \rgtpK}
	\end{axis}

\end{tikzpicture}%
%
%
\begin{tikzpicture}
	
	\begin{axis}[%
	width=1.5in, 
height=2in, 
		at={(1.262in,0.7in)},
		scale only axis,  
		xmin=10,  
		xmax=50,  
		xtick={0, 20,  40,  60}  ,
		xticklabels={{$0$},{$20$}, {$40$},  {$60$}},
		xlabel style={font=\color{white!15!black}},  
		ymin=-8,  
		ymax=1,  
		ytick={0  , -3, -6}  ,
		yticklabels={{} , {}, {}}  ,
		ylabel style={font=\color{white!15!black}}, 
		yminorticks=true,
		axis background/.style={fill=white},
		legend style={at={(.97, .7)}, legend cell align=left, align=left, draw=white!15!black},
		]
		\addplot[color=black, line width=1.0pt, forget plot]
		table[row sep=crcr]{%
		1    0.8936679992604543 \\
		2    0.6377495102237042 \\
		3    0.38440256521497523 \\
		4    0.1338646637378149 \\
		5    -0.11358211904714159 \\
		6    -0.35761270249113275 \\
		7    -0.5978647571227939 \\
		8    -0.8339496575024649 \\
		9    -1.065469227728547 \\
		10    -1.292038073783103 \\
		11    -1.513310135493595 \\
		12    -1.7290067758132937 \\
		13    -1.9389426453521903 \\
		14    -2.1430451834134367 \\
		15    -2.3413642767261758 \\
		16    -2.5340702719289387 \\
		17    -2.7214407785334562 \\
		18    -2.9038387957047513 \\
		19    -3.0816860065674163 \\
		20    -3.2554353190255885 \\
		21    -3.425546029267847 \\
		22    -3.5924637583876065 \\
		23    -3.756606021713411 \\
		24    -3.9183532576802107 \\
		25    -4.078044500093129 \\
		26    -4.235976603217928 \\
		27    -4.392405923472505 \\
		28    -4.547551510217032 \\
		29    -4.701599068363975 \\
		30    -4.854705166981019 \\
		31    -5.007001349605696 \\
		32    -5.158597942857027 \\
		33    -5.309587461091268 \\
		34    -5.460047572638894 \\
		35    -5.61004363539302 \\
		36    -5.759630833458215 \\
		37    -5.908855958154028 \\
		38    -6.057758880361486 \\
		39    -6.206373760182974 \\
		40    -6.354730036289086 \\
		41    -6.502853232551909 \\
		42    -6.65076561445224 \\
		43    -6.79848672280066 \\
		44    -6.946033807781398 \\
		45    -7.093422182340965 \\
		46    -7.240665510495731 \\
		47    -7.387776043242568 \\
		48    -7.534764812325596 \\
		49    -7.6816417901343765 \\
		50    -7.828416022353584 \\
		51    -7.975095738694866 \\
		52    -8.121688445944311 \\
		53    -8.268201006704656 \\
		54    -8.414639706541777 \\
		55    -8.561010311691035 \\
		56    -8.70731811898959 \\
		57    -8.85356799947774 \\
		58    -8.999764436656273 \\
		59    -9.145911560379119 \\
		60    -9.292013176990817 \\
		};
		
		\addplot[color=red!60, dashed, line width=1.0pt,forget plot]
		table[row sep=crcr]{%
		1    0.9125474814223375 \\
		2    0.7032707545022269 \\
		3    0.3953846248502097 \\
		4    0.2020455394487866 \\
		5    -0.12893890803410216 \\
		6    -0.33873641098952995 \\
		7    -0.6575914236246156 \\
		8    -0.8132859032482915 \\
		9    -1.0689005763803472 \\
		10    -1.3599883465429354 \\
		11    -1.6413131310345443 \\
		12    -1.8786366947968292 \\
		13    -2.1343450243783355 \\
		14    -2.5526034540765066 \\
		15    -2.663924112895378 \\
		16    -2.673694942964776 \\
		17    -2.726160276045168 \\
		18    -2.7541748155545585 \\
		19    -2.372531691435045 \\
		20    -2.1887191101853327 \\
		21    -2.7696386840045184 \\
		22    -2.4004624467771225 \\
		23    -2.2711937102722803 \\
		24    -3.0113392103349685 \\
		25    -2.3324717929697916 \\
		26    -2.317711502012684 \\
		27    -3.1993649968616573 \\
		28    -2.2971478782891777 \\
		29    -2.3400039487884614 \\
		30    -3.2844435586817156 \\
		31    -2.2816392299774075 \\
		32    -2.350383196361741 \\
		33    -3.321128923499892 \\
		34    -2.274725465560492 \\
		35    -2.3550744760021143 \\
		36    -3.338118154744007 \\
		37    -2.2715747955811736 \\
		38    -2.3571319804748287 \\
		39    -3.3464857360491096 \\
		40    -2.270133182689367 \\
		41    -2.358000357569272 \\
		42    -3.350752891664128 \\
		43    -2.2694796298806166 \\
		44    -2.3583456033662493 \\
		45    -3.3529757372377635 \\
		46    -2.2691887966883018 \\
		47    -2.358468554502282 \\
		48    -3.354153071289905 \\
		49    -2.2690630149532907 \\
		50    -2.3585020365200857 \\
		51    -3.3547861347014654 \\
		52    -2.2690109621239016 \\
		53    -2.3585028617612784 \\
		54    -3.3551314093215017 \\
		55    -2.2689909673507564 \\
		56    -2.3584941532861454 \\
		57    -3.3553222334710013 \\
		58    -2.268984354116848 \\
		59    -2.3584845351054073 \\
		60    -3.3554289811365967 \\	
		};
		
		\addplot[color=blue!60, dotted, line width=1.0pt,forget plot]
		table[row sep=crcr]{%
		1    0.9125474814223375 \\
		2    0.6821882943151681 \\
		3    0.39121079705251827 \\
		4    0.14967278787957758 \\
		5    -0.10657642091327554 \\
		6    -0.35107161536994713 \\
		7    -0.568042661923196 \\
		8    -0.7605592698008034 \\
		9    -1.1324568986173758 \\
		10    -1.319718945790481 \\
		11    -1.2483238523678972 \\
		12    -1.82806445193756 \\
		13    -2.0072136666413187 \\
		14    -2.2512362607977656 \\
		15    -2.4371894060194186 \\
		16    -2.62347746891707 \\
		17    -2.7988825931506236 \\
		18    -2.973675887469815 \\
		19    -3.1570622354782922 \\
		20    -3.3050121126109833 \\
		21    -3.495635646157324 \\
		22    -3.6321510521596285 \\
		23    -3.823413155140554 \\
		24    -3.94206954571859 \\
		25    -4.146045450231873 \\
		26    -4.269631369932213 \\
		27    -4.46327952558989 \\
		28    -4.564832002960283 \\
		29    -4.780429879058151 \\
		30    -4.8989906052985495 \\
		31    -5.09222847267745 \\
		32    -5.182256520369713 \\
		33    -5.406783913087212 \\
		34    -5.523143643832494 \\
		35    -5.7149204428298415 \\
		36    -5.834609928726715 \\
		37    -6.020784997685 \\
		38    -6.142518086367863 \\
		39    -6.324872659924477 \\
		40    -6.447985711858307 \\
		41    -6.627583751352979 \\
		42    -6.751682995438337 \\
		43    -6.929197023608871 \\
		44    -7.054052796802725 \\
		45    -7.229920197989029 \\
		46    -7.355401225437181 \\
		47    -7.529915717468903 \\
		48    -7.655948266600641 \\
		49    -7.829314531293465 \\
		50    -7.955857223622527 \\
		51    -8.128224019673773 \\
		52    -8.255252534094838 \\
		53    -8.42673296764966 \\
		54    -8.554230986125857 \\
		55    -8.724914995669188 \\
		56    -8.852869059453147 \\
		57    -9.022831133462171 \\
		58    -9.15122791361472 \\
		59    -9.320531874115357 \\
		60    -9.449356898226002 \\
		};
	\end{axis}
\end{tikzpicture}%
		
%
%
\begin{tikzpicture}
	
	\begin{axis}[%
	width=1.5in, 
height=2in, 
		at={(1.262in,0.7in)},
		scale only axis,  
		xmin=10,  
		xmax=50,  
		xtick={0, 20,  40,  60}  ,
		xticklabels={{$0$},{$20$}, {$40$},  {$60$}},
		xlabel style={font=\color{white!15!black}},  
		xlabel={iteration},  
		ymin=-8,  
		ymax=1,  
		ytick={0  , -3, -6}  ,
		yticklabels={{$0$} , {$-3$}, {$-6$}}  ,
		ylabel style={font=\color{white!15!black}}, 
		ylabel={$\log \delta^t$},
		yminorticks=true,
		axis background/.style={fill=white},
		legend style={at={(.97, .7)}, legend cell align=left, align=left, draw=white!15!black},
		]
		\addplot[color=black, line width=1.0pt]
		table[row sep=crcr]{%
		1    0.0009784296723861028 \\
		2    -0.3968502364829041 \\
		3    -0.7726955140154359 \\
		4    -1.1170649077521595 \\
		5    -1.4218114016443466 \\
		6    -1.684119189485988 \\
		7    -1.9081252757962743 \\
		8    -2.102436195659814 \\
		9    -2.2761515623461714 \\
		10    -2.4364633432108307 \\
		11    -2.5882173775620014 \\
		12    -2.734409816708569 \\
		13    -2.876821447347146 \\
		14    -3.0165024635539495 \\
		15    -3.1540824498684827 \\
		16    -3.2899530576753517 \\
		17    -3.4243715257656637 \\
		18    -3.557518345690491 \\
		19    -3.6895292340760752 \\
		20    -3.8205129105558595 \\
		21    -3.950561057531086 \\
		22    -4.079753957648678 \\
		23    -4.208163723602705 \\
		24    -4.335856174993631 \\
		25    -4.462891949459571 \\
		26    -4.589327179318905 \\
		27    -4.715213923022196 \\
		28    -4.840600460658968 \\
		29    -4.965531516744826 \\
		30    -5.090048446566858 \\
		31    -5.214189406352771 \\
		32    -5.337989517949569 \\
		33    -5.461481032987822 \\
		34    -5.584693498175256 \\
		35    -5.707653921504367 \\
		36    -5.83038693821595 \\
		37    -5.952914974978733 \\
		38    -6.075258410698462 \\
		39    -6.197435732509386 \\
		40    -6.31946368573655 \\
		41    -6.441357416884801 \\
		42    -6.563130608980639 \\
		43    -6.684795608837887 \\
		44    -6.806363546035268 \\
		45    -6.927844443574072 \\
		46    -7.049247320330586 \\
		47    -7.170580285529004 \\
		48    -7.291850625548051 \\
		49    -7.413064883432026 \\
		50    -7.534228931516504 \\
		51    -7.65534803760153 \\
		52    -7.776426925114303 \\
		53    -7.897469827701207 \\
		54    -8.018480538679505 \\
		55    -8.139462455764791 \\
		56    -8.260418621471096 \\
		57    -8.38135175955636 \\
		58    -8.502264307869504 \\
		59    -8.623158447920073 \\
		60    -8.744036131483732 \\
		};
		
		\addplot[color=red!60, dashed, line width=1.0pt]
		table[row sep=crcr]{%
		1    0.023196851377182453 \\
		2    -0.3392881209797888 \\
		3    -0.4779180257485619 \\
		4    -1.2324806393455014 \\
		5    -1.2723717503027399 \\
		6    -1.5703816770282302 \\
		7    -1.4922157235174736 \\
		8    -1.693895697924828 \\
		9    -1.5487040477286902 \\
		10    -2.0255738846552354 \\
		11    -1.9703839187088223 \\
		12    -2.754962575114595 \\
		13    -2.1793694908492967 \\
		14    -2.663058689346279 \\
		15    -2.4540633264190856 \\
		16    -2.5632349148328575 \\
		17    -2.5920034716571063 \\
		18    -2.5199178884955495 \\
		19    -2.649264748261775 \\
		20    -2.5040860733428403 \\
		21    -2.669061628411616 \\
		22    -1.8658413061095351 \\
		23    -2.306105120798349 \\
		24    -2.4237190340560213 \\
		25    -2.159150691035807 \\
		26    -2.4169774021203616 \\
		27    -2.17234283241893 \\
		28    -2.087053139409115 \\
		29    -2.3238474296107614 \\
		30    -2.8537275933835846 \\
		31    -2.44391346022792 \\
		32    -2.673265930529994 \\
		33    -2.569487647213843 \\
		34    -2.603064584617977 \\
		35    -2.6268438087245736 \\
		36    -2.5746798042446986 \\
		37    -2.6515361550453918 \\
		38    -2.5630368947023574 \\
		39    -2.6618750426463884 \\
		40    -2.558253363785402 \\
		41    -2.6661143899454327 \\
		42    -2.5563062927120814 \\
		43    -2.6678062101527447 \\
		44    -1.8636653901086462 \\
		45    -2.26102878464488 \\
		46    -2.399719797930534 \\
		47    -2.1263515541541778 \\
		48    -2.0949934307435667 \\
		49    -2.2851775583768728 \\
		50    -2.8048290105217797 \\
		51    -2.421184238839982 \\
		52    -2.6209780176343584 \\
		53    -2.550506716469651 \\
		54    -2.5488394423443275 \\
		55    -2.6095276505585017 \\
		56    -2.519417571538645 \\
		57    -2.6352025967538326 \\
		58    -2.507141998107553 \\
		59    -2.6462230100278377 \\
		60    -2.5019485234630765 \\
		};
		
		\addplot[color=blue!60, dotted, line width=1.0pt]
		table[row sep=crcr]{%
		1    0.023196851377182453 \\
		2    -0.3392881209797888 \\
		3    -0.42595827141693215 \\
		4    -1.3272891478229973 \\
		5    -1.3838343004425921 \\
		6    -1.7990687044534495 \\
		7    -1.8558886900219318 \\
		8    -2.134987311177401 \\
		9    -2.193477656378168 \\
		10    -2.5309889485468955 \\
		11    -2.431787509985634 \\
		12    -2.5456420303574805 \\
		13    -2.07186338481453 \\
		14    -3.1360768921503195 \\
		15    -3.117411370475079 \\
		16    -3.3963378217625224 \\
		17    -3.4211399111276735 \\
		18    -3.6855652016660603 \\
		19    -3.730057383109466 \\
		20    -3.93570187955735 \\
		21    -4.006254569384792 \\
		22    -4.189715664527044 \\
		23    -4.272870782525625 \\
		24    -4.446896203288548 \\
		25    -4.535994313249414 \\
		26    -4.7053524810640335 \\
		27    -4.797469586185183 \\
		28    -4.964389429766538 \\
		29    -5.05803866260188 \\
		30    -5.2237020701181756 \\
		31    -5.318074184445713 \\
		32    -5.4831314802040945 \\
		33    -5.5777857467937615 \\
		34    -5.742593608508757 \\
		35    -5.837296726859753 \\
		36    -6.002046707435135 \\
		37    -6.096681449356174 \\
		38    -6.261472494676836 \\
		39    -6.355985528902679 \\
		40    -6.520865096364477 \\
		41    -6.6152374767779385 \\
		42    -6.780224784363709 \\
		43    -6.874455442665081 \\
		44    -7.03955456220427 \\
		45    -7.133651185860016 \\
		46    -7.298858378597789 \\
		47    -7.392832449081726 \\
		48    -7.558140243115671 \\
		49    -7.652004400601489 \\
		50    -7.817403828183386 \\
		51    -7.911170526171886 \\
		52    -8.076652324596001 \\
		53    -8.17033319146773 \\
		54    -8.33588842281532 \\
		55    -8.429494004072087 \\
		56    -8.595114351474983 \\
		57    -8.688654051302105 \\
		58    -8.854331937359259 \\
		59    -8.947814059591305 \\
		60    -9.113542669091885 \\
		};
		
	\end{axis}
\end{tikzpicture}%
%
%
\begin{tikzpicture}
	
	\begin{axis}[%
	width=1.5in, 
height=2in, 
		at={(1.262in,0.7in)},
		scale only axis,  
		xmin=10,  
		xmax=50,  
		xtick={0, 20,  40,  60}  ,
		xticklabels={{$0$},{$20$}, {$40$},  {$60$}},
		xlabel style={font=\color{white!15!black}},  
		xlabel={iteration},  
		ymin=-8,  
		ymax=1,  
		ytick={0  , -3, -6}  ,
		yticklabels={{} , {}, {}}  ,
		ylabel style={font=\color{white!15!black}}, 
		yminorticks=true,
		axis background/.style={fill=white},
		legend style={at={(.97, .7)}, legend cell align=left, align=left, draw=white!15!black}, 
		]
		\addplot[color=black, line width=1.0pt]
		table[row sep=crcr]{%
		1    0.08449300474126609 \\
		2    -0.12865183616699427 \\
		3    -0.32719752447468126 \\
		4    -0.5141843668426168 \\
		5    -0.6924441268939008 \\
		6    -0.8643266347284397 \\
		7    -1.0316367939139444 \\
		8    -1.195690836495518 \\
		9    -1.357416243430846 \\
		10    -1.5174524164576673 \\
		11    -1.6762345609741183 \\
		12    -1.8340572294672712 \\
		13    -1.9911198899651337 \\
		14    -2.1475585216298647 \\
		15    -2.3034670546773803 \\
		16    -2.458911712774272 \\
		17    -2.6139405139378296 \\
		18    -2.7685895199511754 \\
		19    -2.922886924698479 \\
		20    -3.0768557164815986 \\
		21    -3.230515404337359 \\
		22    -3.3838831325808276 \\
		23    -3.5369743969832568 \\
		24    -3.6898035025172984 \\
		25    -3.842383854150269 \\
		26    -3.9947281403381605 \\
		27    -4.146848448023341 \\
		28    -4.298756334309156 \\
		29    -4.450462871089658 \\
		30    -4.601978673116085 \\
		31    -4.753313916209287 \\
		32    -4.904478349876123 \\
		33    -5.055481306997985 \\
		34    -5.2063317122312815 \\
		35    -5.357038090096971 \\
		36    -5.507608573311842 \\
		37    -5.658050911644329 \\
		38    -5.80837248140926 \\
		39    -5.958580295612301 \\
		40    -6.1086810146942785 \\
		41    -6.258680957791719 \\
		42    -6.40858611441403 \\
		43    -6.558402156432595 \\
		44    -6.708134450279849 \\
		45    -6.857788069260556 \\
		46    -7.00736780588774 \\
		47    -7.156878184163384 \\
		48    -7.30632347173581 \\
		49    -7.455707691872113 \\
		50    -7.605034635196507 \\
		51    -7.754307871150735 \\
		52    -7.9035307591458395 \\
		53    -8.052706459371624 \\
		54    -8.20183794324846 \\
		55    -8.350928003505604 \\
		56    -8.499979263873364 \\
		57    -8.648994188379156 \\
		58    -8.79797509025918 \\
		59    -8.946924140470518 \\
		60    -9.095843375813152 \\
		};
		
		\addplot[color=red!60, dashed, line width=1.0pt]
		table[row sep=crcr]{%
		1    0.17027225875564272 \\
		2    0.039385472969830304 \\
		3    -0.21388524066296571 \\
		4    -0.2524708399718872 \\
		5    -0.5380737727276864 \\
		6    -0.6007498411274718 \\
		7    -0.7630681088819372 \\
		8    -0.8071154319263854 \\
		9    -1.1404303163348386 \\
		10    -1.0880941983445518 \\
		11    -1.1747117600137151 \\
		12    -1.3328271443278923 \\
		13    -1.8162084119517339 \\
		14    -1.3209822667748177 \\
		15    -1.5162098138761677 \\
		16    -2.3734343872612977 \\
		17    -2.374728446533494 \\
		18    -2.3754537298942053 \\
		19    -2.3579950558957687 \\
		20    -2.3585834481456405 \\
		21    -1.4512363584691945 \\
		22    -1.2709588975518873 \\
		23    -1.4908734484483959 \\
		24    -1.2868480805949623 \\
		25    -1.623961014346534 \\
		26    -1.293054727323429 \\
		27    -1.2896010396029456 \\
		28    -1.5085261351854955 \\
		29    -1.7670355157076263 \\
		30    -1.4399417259235554 \\
		31    -1.270873441929732 \\
		32    -1.811069388575741 \\
		33    -1.8079587102164163 \\
		34    -1.5769963822281525 \\
		35    -1.581272848530222 \\
		36    -1.3238754018286274 \\
		37    -1.6088851692361843 \\
		38    -1.839435264421566 \\
		39    -1.8651296048611616 \\
		40    -1.5238421770880404 \\
		41    -1.5264133314796933 \\
		42    -1.3593446453020044 \\
		43    -1.6234056468680147 \\
		44    -1.638549629626684 \\
		45    -1.2418242656617504 \\
		46    -1.4038392785361637 \\
		47    -1.4191166270935134 \\
		48    -1.6860408788365855 \\
		49    -1.3494116193143997 \\
		50    -1.2575583401245354 \\
		51    -1.7064789523106914 \\
		52    -1.7039797102050522 \\
		53    -1.605349647631346 \\
		54    -1.6113601886545965 \\
		55    -1.3850675379268866 \\
		56    -1.723744688828458 \\
		57    -1.488805084354476 \\
		58    -1.29536604260953 \\
		59    -1.651563000370936 \\
		60    -1.2859655125095566 \\
		};
		
		\addplot[color=blue!60, dotted, line width=1.0pt]
		table[row sep=crcr]{%
		1    0.17027225875564272 \\
		2    0.039385472969830304 \\
		3    -0.21388524066296571 \\
		4    -0.2524708399718872 \\
		5    -0.5380737727276864 \\
		6    -0.6007498411274718 \\
		7    -0.8771330589415883 \\
		8    -0.8981379589479486 \\
		9    -1.1153481300109447 \\
		10    -1.1007271679403585 \\
		11    -1.154046098304124 \\
		12    -1.1250786861346327 \\
		13    -1.3197215436323406 \\
		14    -1.6064657815324204 \\
		15    -1.439028020337428 \\
		16    -1.6908854644182059 \\
		17    -2.5961689547081006 \\
		18    -1.5577579068042389 \\
		19    -1.4828207500116768 \\
		20    -2.4808086209110125 \\
		21    -2.48038775422053 \\
		22    -3.4007873206597914 \\
		23    -3.400458676583313 \\
		24    -1.6104970617887888 \\
		25    -1.6095784577349528 \\
		26    -1.906280213282132 \\
		27    -1.9067600410407246 \\
		28    -2.040094065594796 \\
		29    -2.0410330967744343 \\
		30    -2.2133062022773164 \\
		31    -2.2066362980191014 \\
		32    -1.7091898721418972 \\
		33    -1.7055337067322482 \\
		34    -1.7157809843103224 \\
		35    -1.7176086603431124 \\
		36    -2.1357918751924725 \\
		37    -2.145754808602592 \\
		38    -2.1407775252994448 \\
		39    -2.1420234555138133 \\
		40    -2.150706863736443 \\
		41    -2.151593799996278 \\
		42    -1.806477761680899 \\
		43    -1.8039486723926563 \\
		44    -1.85659300462944 \\
		45    -1.8552206412159176 \\
		46    -2.288580491884174 \\
		47    -2.2957092693446675 \\
		48    -2.298561308437371 \\
		49    -2.296099515848311 \\
		50    -1.942758744133276 \\
		51    -1.94237480883198 \\
		52    -1.9467012741468588 \\
		53    -1.9469199753945317 \\
		54    -1.9272220296434142 \\
		55    -1.9303330282778919 \\
		56    -2.2555574013004454 \\
		57    -2.2488369653023335 \\
		58    -2.3016085772249406 \\
		59    -2.301455383099788 \\
		60    -1.7991808798042754 \\
		};
		
	\end{axis}
\end{tikzpicture}%
	\end{center}\vspace{7mm}
	\caption{
		Optimality gap against the number of iterations in the low-dimensional case with $J=4$ for all possible sparsification factors in a sample problem: $S=1$ (top left), $S=0.75$ (top right), $S=0.5$ (bottom left), and $S=0.25$ (bottom right). Even in the low-dimensional setting \rgtpK can significantly outperform \tpK.}
	\label{fig:low-dim}
\end{figure*}

\section{Closer Look at \rgtpK in Lower Dimensions}
\label{append:A}
This appendix gives a quantitative illustration for the intuitive discussions given in Section~\ref{sec:Num}. To this end, we consider a sample low-dimensional example in which we can track the error accumulation numerically. We consider the same setting as in the linear regression problem investigated in Section~\ref{sec:Num} with $N=2$ workers and $J=4$ model parameters. We keep the ratio of the data points to model size similar to that considered in the previous experiment, i.e., we set $D_n = 20$ for both workers. The datasets are generated according to the mechanism illustrated  in Section~\ref{sec:linReg} with mean $U=0$, variances $\sigma^2 = h^2 = 1$, and error variance $\epsilon^2 = 0.5$.

\subsection{Convergence to Global Optimum}
For this setting, we have only four possible sparsity factors, namely $S=1$, $S=0.75$, $S=0.5$, and $S=0.25$, which correspond to the cases of $k=4$ (no sparsification), $k=3$, $k=2$, and $k=1$, respectively. For each of these cases, the distance $\delta^t$ is plotted against the number of iterations in Figure~\ref{fig:low-dim}. As the figure shows, in this low dimensional example \tpK can never converge to the global optimum: for any $S\neq 1$, \tpK remains somewhere at a fixed distance from the global optimum. In contrast, \rgtpK converges to the global optimum at any $S\neq 0.25$.

\begin{table*}[h!]
	\centering
	\caption{Accumulated gradients for \tpK and \rgtpK in the case of $S=0.75$: \normalfont Bold entry shows the entry corresponding to the largest aggregated gradient in non-sparsified case, i.e., global largest. After iteration 23, \tpK removes frequently this entry at the workers, whereas \rgtpK can keep it. }
		\begin{tabular}[t]{cccc}
	\hline
	\vspace{-3mm}\\
	iteration &Aggregation Target &\textsc{Top-}$k$ &\textsc{RegTop-}$k$
	\vspace{1mm}\\
	\hline
	\vspace{-3mm}\\
	1
	&$\begin{bmatrix}
		-6.6994\\ -5.6918\\ -4.2025\\ -0.5833
	\end{bmatrix} $,
	&$\begin{bmatrix}
		-6.25\\	-4.67\\0.00\\2.76
	\end{bmatrix} $,
	$\begin{bmatrix}
		-7.14\\ -6.70\\ -7.46\\ 0.00
	\end{bmatrix} $
	&$\begin{bmatrix}
		-6.25\\	-4.67\\0.00\\2.76
	\end{bmatrix} $,
	$\begin{bmatrix}
		-7.14\\ -6.70\\ -7.46\\ 0.00
	\end{bmatrix} $
	\\
	$\vdots$ &$\vdots$ &$\vdots$ &$\vdots$\\
\vspace{-3mm}\\
		23
	&$\begin{bmatrix}
		\mathbf{-0.0221}\\ -0.0044\\  0.0063\\ -0.0155\\
	\end{bmatrix}$ 
	&$\begin{bmatrix}
		\mathbf{1.14}\\ 0.00\\ 1.48\\ 2.49
	\end{bmatrix} $,
	$\begin{bmatrix}
		\mathbf{0.00}\\ -2.46\\ -1.49\\ -2.52
	\end{bmatrix} $
	&$\begin{bmatrix}
		\mathbf{0.39}\\ 0.00\\ 1.49\\ 4.98\\
	\end{bmatrix} $,
	$\begin{bmatrix}
		\mathbf{-0.43}\\  0.00\\ -1.48\\ -5.03
	\end{bmatrix} $
	\\
	\vspace{-3mm}\\
			24
	&$\begin{bmatrix}
		\mathbf{-0.0178}\\ -0.0032\\  0.0061\\ -0.0126\\
	\end{bmatrix}$ 
	&$\begin{bmatrix}
		\mathbf{0.00}\\ 2.37\\ 1.50\\ 2.48\\
	\end{bmatrix} $,
	$\begin{bmatrix}
		\mathbf{0.00}\\ -1.05\\ -1.46\\ -2.51\\
	\end{bmatrix} $
	&$\begin{bmatrix}
		\mathbf{0.40}\\ 2.34\\ 0.00\\ 2.50\\
	\end{bmatrix} $,
	$\begin{bmatrix}
		\mathbf{-0.43}\\ -2.34\\  0.00\\ -2.51\\
	\end{bmatrix} $
	\\
	$\vdots$ &$\vdots$ &$\vdots$ &$\vdots$\\
	\vspace{-3mm}\\
				40
	&$\begin{bmatrix}
	-0.0007\\ -0.0001\\  	\mathbf{0.0008}\\ -0.0006\\
	\end{bmatrix}$ 
	&$\begin{bmatrix}
		0.00\\ 1.16\\ \mathbf{1.48}\\ 2.52\\
	\end{bmatrix} $,
	$\begin{bmatrix}
	-1.25\\  0.00\\ \mathbf{-1.45}\\ -2.50\\
	\end{bmatrix} $
	&$\begin{bmatrix}
		0.41\\ 2.34\\ \mathbf{1.48}\\ 0.00
	\end{bmatrix} $,
	$\begin{bmatrix}
	-0.41\\ -2.34\\ \mathbf{-1.48}\\  0.00
	\end{bmatrix} $\vspace{1mm}
	\\
	\hline
\end{tabular}
	\label{table:1}
\end{table*}%

\subsection{Tracking Error Accumulation}
We now track the accumulated gradients of the workers for the case $S=0.75$ considering both algorithms. It is worth recalling that in this case $k=3$, and hence each worker drops only one of its gradient entries in each iteration. The accumulated gradients for some sample iterations are shown in Table~\ref{table:1}. In this table, we record the sparsified gradients sent by the workers, as well as the \textit{aggregation target}, i.e., the aggregated gradient without sparsification. For sake of compactness, the entries of local gradients are rounded to two decimal points.

The bold entry in the table corresponds to the largest \textit{aggregated gradient entry}. This means, in the genie-aided case of global \tpK, this entry \textit{should not} be dropped at workers. Let us now, take a look at the local gradients before sparsification: in the first iteration, \tpK and \rgtpK determine the same gradients. This is due to the fact that \rgtpK has no prior information in the first iteration and simply performs \tpK for sparsification. We now let the algorithm iterate and go to iteration 23, where as Figure~\ref{fig:low-dim} shows, \tpK fails to move further towards the global optimum. Comparing the local gradients of \tpK and \rgtpK with the target aggregation at this iteration, one can see that \tpK drops the first entry at the second worker, which is corresponding to the largest aggregated entry in the non-sparsified case. Unlike \tpK, \rgtpK is able to keep the entry that is corresponding to the largest aggregated value. This behavior is repeated again in iteration 24, where using \tpK leads both workers to drop the local entry whose aggregation would be the largest. As we reach iteration 40, one can observe that using \tpK, the server aggregates a gradient that is considerably different from non-sparsified aggregation, while \rgtpK tracks the aggregation target rather closely. This latter observation quantitatively describes our initial intuitive discussions on learning rate scaling.

\subsection{Comparing Sparsification Masks}
There is another interesting aspect seen in our numerical investigations that is worth mentioning: as \rgtpK evolves, one can observe high overlap between the sparsification masks across the workers. For instance in Table~\ref{table:1}, using \rgtpK the workers drop the same entry, although they sparsify their accumulated gradients independently. This is due to the fact that \rgtpK utilizes the common information, i.e., previous aggregated gradients, for sparsification. In other words, using the information that is shared among the workers, \rgtpK manages to implicitly coordinate the sparsification masks across the workers.

\end{document}